\documentclass[letterpaper]{article}

\usepackage{aaai}

\usepackage{times}
\usepackage{helvet}
\usepackage{courier}
\usepackage{standalone}
\usepackage[algo2e,ruled,noline,linesnumbered]{algorithm2e}
\makeatletter
\newcommand{\removelatexerror}{\let\@latex@error\@gobble}
\makeatother
\usepackage[usenames,dvipsnames]{color}
\usepackage{tikz}
\usetikzlibrary{shapes,arrows}
\usetikzlibrary{positioning}
\usetikzlibrary{chains}
\usetikzlibrary{trees}
\usepackage{amssymb}
\usepackage{amsmath}
\usepackage{amsthm}
\usepackage{subfigure}
\usepackage{url}

\newcommand{\argmax}{\operatornamewithlimits{argmax}}
\DeclareMathOperator{\mP}{\mathcal{P}}

\DeclareMathOperator{\mD}{\mathcal{D}}

\DeclareMathOperator{\E}{\mathbb{E}}

\DeclareMathAlphabet\mathbfcal{OMS}{cmsy}{b}{n}

\newcommand*\xbar[1]{%
  \hbox{%
    \vbox{%
      \hrule height 0.5pt 
      \kern0.5ex
      \hbox{%
        \kern-0.1em
        \ensuremath{#1}%
        \kern-0.1em
      }%
    }%
  }%
}

\usepackage[normalem]{ulem}
\definecolor{RED}{rgb}{1,0,0}\definecolor{BLUE}{rgb}{0,0,1}

\providecommand{\red}[1]{{\textcolor{red}{#1}}}

\usepackage{tikz}
\usepackage{dsfont}
\begin{document}
\title{Better Optimism By Bayes: Adaptive Planning with Rich Models}
\author{Arthur Guez\\Gatsby Unit, UCL\\aguez@gatsby.ucl.ac.uk \And David Silver\\ Dept. of Computer Science, UCL\\ d.silver@cs.ucl.ac.uk\\ \And Peter Dayan\\ Gatsby Unit, UCL\\ dayan@gatsby.ucl.ac.uk}
\maketitle

\begin{abstract}
\begin{quote}
The computational costs of inference and planning have confined Bayesian
model-based reinforcement learning to one of two dismal fates: powerful
Bayes-adaptive planning but only for simplistic models, or powerful,
Bayesian non-parametric models but using simple, myopic planning
strategies such as Thompson sampling.  We ask whether it is feasible and
truly beneficial to combine rich probabilistic models with a closer
approximation to fully Bayesian planning.

First, we use a collection of counterexamples to show formal problems
with the over-optimism inherent in Thompson sampling. Then we leverage
state-of-the-art techniques in efficient Bayes-adaptive planning and
non-parametric Bayesian methods to perform qualitatively better than
both existing conventional algorithms and Thompson sampling on two
contextual bandit-like problems.
\end{quote}
\end{abstract}

\noindent As computer power increases and statistical methods improve,
there is an increasingly rich range and variety of probabilistic models
of the world.  Models embody inductive biases, allowing appropriately
confident inferences to be drawn from limited observations. One domain
that should benefit markedly from such models is planning and control
--- models arbitrate the exquisite balance between safe exploration and
lucrative exploitation.

A general and powerful solution to this balancing act involves
forward-looking Bayesian planning in the face of partial observability,
which treats the exploration-exploitation trade-off as an {\it
optimization\/} problem, squeezing the greatest benefit from each
choice. Unfortunately, this is notoriously computationally costly,
particularly for complex models, leaving open the possibility that it
might not be justified compared to heuristic approaches that may perform
very similarly at a much reduced computational cost, for instance
treating the tradeoff as a {\em learning} problem
in a regret setting, focusing on an asymptotic requirement to
discover the optimal solution (to avoid accumulating regret).

The motivation for this paper is to demonstrate the practical power of
Bayesian planning. We show that, despite the arduous
optimization problem, sample-based planning approximations can excel
with rich models in realistic settings -- here a challenging
exploration-exploitation task derived from a real dataset (the UCI
'mushroom' task) -- \textit{even when the data have not been generated
from the prior.}  By contrast, we show that the benefits of 
Bayesian inference can be squandered by more myopic forms of planning
--- such as the provably over-optimistic Thompson Sampling -- which 
fails to account for risk in this task and performs poorly.  The
experimental results highlight the fact that the Bayes-optimal behavior
adapts its exploration strategy as a function of the cost, the horizon,
and the uncertainty in a non-trivial way. We also consider an extension
of the model to a case of more general subtasks, including subtasks that
are themselves small {\sc mdp}s (in the suppl. material, Section~\ref{sec:crpmdp}).

The paper is organized as follows: first, we discuss model-based
Bayesian reinforcement learning (RL), outline some existing planning
algorithms for this case and show why Thompson sampling's over-optimism
can be deleterious.  Next, we introduce an exploration-exploitation
domain that motivates a statistical model for a class of {\sc mdp}s with
shared structure across sequences of tasks. We provide empirical results
on a version of the domain that uses real data coming from a popular
supervised learning problem (mushroom classification) along with a
simulated extension. Finally, we discuss related work.

\section{Model-based Bayesian RL}

We consider a Bayes-Adaptive planner~\cite{Duff:2002:aaa}, which starts
with a prior $P(\mP)$ over models of the environmental dynamics,
progressively receives data $\mD$ through controlled interaction with
the environment, and updates its posterior distribution over models
using Bayes-rule $P(\mP | \mD) \propto P(\mD | \mP) P(\mP)$.  Actions are
intended to maximize an expected discounted return criterion
$\E[\sum_{t=0}^\infty \gamma^t r_t]$, where $\gamma<1$ is the discount
factor and $r_t$ is the random reward obtained at time $t$. In an
uncertain world, this requires balancing exploration and exploitation.
Here, the discount factor $\gamma$ plays the crucial role of arbitrating
the relative importance of future rewards. In general, a low $\gamma$
does not warrant much exploration because future exploitation will be
heavily downweighted. The opposite is true as $\gamma
\rightarrow 1$.  A clear illustration of these $\gamma$-dependent
exploration-exploitation Bayesian policies can be found in the Gittins
indices~\cite{Gittins:1989:aaa}.
\footnote{Even though we refer to exploration and
exploitation, actions are never actually labeled with one or the other
in this Bayesian setting, it is only an interpretation for actions whose
consequences are more uncertain (explore) or more certainly valuable
(exploit).}

The (Bayes-)optimal strategy integrates over how the current belief
$P(\mP | \mD)$ could be transformed in the light of (imaginary)
possible future data. The resulting policy is well known to be the
solution to an augmented Markov Decision Process ({\sc mdp}), whose details
we defer to the suppl.\  material, Section~\ref{sec:bap}. Finding the exact Bayes-optimal
policy is computationally intractable even for tiny state spaces,
since (a) the augmented state space is either continuous or discrete
and potentially unbounded; and (b) the
transitions of the augmented {\sc mdp} require integration over the full
posterior.  Although this operation can be trivial and closed-form for
some simple probabilistic models (e.g., independent
Dirichlet-Multinomial), it is intractable for most rich models.

A common solution to (b) is to use approximate inference methods, such
as Markov chain Monte Carlo ({\sc mcmc}). This fits snugly with a common
heuristic for (a), in which the planning problem is side-stepped by
sampling from the posterior but only planning myopically. We describe
one such method called Thompson Sampling in the section below, but
show that it is no panacea.

A potentially more powerful class of approximate solutions to (a) that
should be capable of handling large state spaces and complex models
involves sample-based forward-search methods. Algorithms such as Sparse
Sampling~\cite{Wang:2005:aaa} or {\sc bamcp}~\cite{Guez:2012:aaa} do not
plan myopically; they approximate Bayes-adaptive planning directly ---
albeit at a computational cost. It had been unclear how to integrate
these methods with approximate, {\sc mcmc}, approaches to (b). However,
recent algorithmic developments \cite{Guez:2013:aaa} provide a practical
way to use approximate inference schemes to perform sample-based
planning with sophisticated models.

\subsection{Thompson Sampling}

Thompson Sampling ({\sc ts})~\cite{Thompson:1933:aaa} is a
myopic planning method that selects actions at each step by 1) drawing a single sample of
the dynamics from the posterior distribution $P(\mP | \mD)$ ; 2)
greedily solving the corresponding sampled {\sc mdp}; and 3) choosing the optimal
action of this {\sc mdp} at the current state. Though
heuristically myopic from the perspective of Bayes-adaptivity, {\sc ts} is
computationally cheap, and has been proven both
empirically~\cite{Chapelle:2011:aaa} and theoretically to perform well in
various domains (including reaching theoretical regret lower-bounds for
multi-armed bandits~\cite{Agrawal:2011:aaa}). As mentioned, it fits well with
complex, e.g., Bayesian non-parametric, models that in any case are handled via
{\sc mcmc} sampling~\cite{Doshi-Velez:2010:aaa}.

Intuitively, {\sc ts} generates optimistic values in unknown parts of the {\sc mdp}
where the posterior entropy over its samples is large. This forces the
agent to visit these regions. However, to show that this way of deriving
optimism for exploration is not always beneficial, we consider two simple,
and yet particularly pernicious, classes of counter-example; other
failure modes are illustrated in the results section below.

\begin{figure}[htb]
\begin{center}
\begin{tikzpicture}[scale=0.8]
    \node (X) at (0,0) {};
    \node (Y) at (6,0) {};
    \draw [semithick] (X) -- (Y);
    \node[draw,circle,thick,inner sep=0pt,minimum size=0.2cm,fill=green] at (0,0) {};
    \node[draw,circle,thick,inner sep=0pt,minimum size=0.2cm,fill=white!30] at (1,0) {};
    \node[draw,circle,thick,inner sep=0pt,minimum size=0.2cm,fill=white!30] at (2,0) {};
    \node[draw,circle,thick,inner sep=0pt,minimum size=0.2cm,fill=blue!30] at (3,0) {};
    \node[draw,circle,thick,inner sep=0pt,minimum size=0.2cm,fill=white!30] at (4,0) {};
    \node[draw,circle,thick,inner sep=0pt,minimum size=0.2cm,fill=white!30] at (5,0) {};
    \node[draw,circle,thick,inner sep=0pt,minimum size=0.2cm,fill=white!30] at (6,0) {};
    \node[draw=none] at (7,0) {w.p $\;1/2$};
    \coordinate (A) at (2,0.15);
    \coordinate (B) at (3,0.1);
    \coordinate (C) at (4,0.15);
    \draw [thick,black,->]   (B) to[out=90,in=90] (A);
    \draw [thick,black,->]   (B) to[out=90,in=90] (C);
    
    \node (X) at (0,-0.5) {};
    \node (Y) at (6,-0.5) {};
    \draw [semithick] (X) -- (Y);
    \node[draw,circle,thick,inner sep=0pt,minimum size=0.2cm,fill=white!30] at (0,-0.5) {};
    \node[draw,circle,thick,inner sep=0pt,minimum size=0.2cm,fill=white!30] at (1,-0.5) {};
    \node[draw,circle,thick,inner sep=0pt,minimum size=0.2cm,fill=white!30] at (2,-0.5) {};
    \node[draw,circle,thick,inner sep=0pt,minimum size=0.2cm,fill=blue!30] at (3,-0.5) {};
    \node[draw,circle,thick,inner sep=0pt,minimum size=0.2cm,fill=white!30] at (4,-0.5) {};
    \node[draw,circle,thick,inner sep=0pt,minimum size=0.2cm,fill=white!30] at (5,-0.5) {};
    \node[draw,circle,thick,inner sep=0pt,minimum size=0.2cm,fill=green] at (6,-0.5) {};
    \node[draw=none] at (7,-0.5) {w.p $\;1/2$};
    \coordinate (D) at (2,-0.35);
    \coordinate (E) at (3,-0.4);
    \coordinate (F) at (4,-0.35);
    \draw [thick,black,->]   (E) to[out=90,in=90] (D);
    \draw [thick,black,->]   (E) to[out=90,in=90] (F);
\end{tikzpicture}
\vspace{-0.1in}
\caption{\footnotesize{Illustration of Example 1. The two possible chains with
$x=3$, with green dots representing the reward states and the blue dot
representing the start state.}}
\label{fig:ex1}
\end{center}
\end{figure}
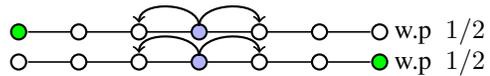

\paragraph{Example 1} \textit{Consider an {\sc mdp} that involves a linear chain of
$2x+1$ states. Each interior state admits $2$ deterministic actions:
going left or right.  The only source of reward ($r=1$) is at either
one or other end. The agent starts in the middle (state $x+1$), and
knows everything except the end which delivers the reward; each of the
two {\sc mdp}s $\mP$ has prior probability $P(\mP)=\frac{1}{2}$. The
episode terminates after the reward is obtained. See Figure~\ref{fig:ex1}
for an illustration.
Critically, the only transition that changes this belief is at an end.
At each step, {\sc ts} samples one of the chains, and so heads for the end
which that sample suggests is rewarding. Since this depends on an
unbiased coin flip, {\sc ts} is effectively performing a random walk with
probability $\frac{1}{2}$ of moving in either direction, and so takes
$O(x^2)$ time to reach an end~\cite{Moon:1973:aaa}. This is much worse
than the linear time of the Bayes-optimal policy which commits to a
given direction by tie-breaking in the first step and then maintains
this direction to the end of the chain.}

One might ascribe this failure to the fact that {\sc ts} was developed for
multi-armed bandits, which lack temporally extended structure. {\sc ts} has
duly been adapted to the {\sc mdp} setting with the
goal of controlling the expected regret.  For instance, the
PSRL algorithm~\cite{Osband:2013:aaa}, which was inspired by 
Bayesian DP~\cite{Strens:2000:aaa}, samples an {\sc mdp} from the
current posterior and executes its optimal policy for several steps
(or an entire episode). This way of exploring an {\sc mdp} bypasses the
{\sc ts}'s lack of commitment in Example 1, but can still be problematic
for discounted objectives, as illustrated in Example 2 (Supp. material).

The BOSS~\cite{Asmuth:2009:aaa} algorithm is a more complicated
construction that combines multiple posterior samples, 
Examples 3-4 (Supp. material) illustrate a similar issue
with the kind of optimism it generates for exploration.

\subsection{Non-myopic planning: Forward-search}
\label{sec:bamcp}

Bayesian planning avoids myopia by integrating over the evolution of
possible future beliefs. Sample-based forward-search planning algorithms
such as Sparse Sampling~\cite{Wang:2005:aaa} perform such integrations,
but they are generally not able to deal with approximate inference
schemes that are necessary to handle rich probabilistic models.

The Bayes-Adaptive Monte-Carlo Planning ({\sc bamcp}) algorithm is a
forward-search, sample-based Bayes-adaptive planning algorithm based on
{\sc pomcp} \cite{Silver:2010:aab} that is guaranteed to converge to the
Bayes-optimal solution, even when combined with {\sc MCMC}-based
inference~\cite{Guez:2012:aaa,Guez:2013:aaa}. Despite its lack of
finite-time guarantees, it displays good empirical performance on a
number of tasks. {\sc bamcp} compounds the advantages of
sparse-sampling~\cite{Wang:2005:aaa} and UCT~\cite{Kocsis:2006:aaa} to
increase search efficiency. It shares with {\sc ts} the use of samples
taken from the posterior; but combines many samples in a search tree to
be able to plan less myopically. Critically, like {\sc pomcp}, {\sc
bamcp} involves root sampling, in which samples are only generated for
the current history from the distribution $P(\mP | h_t)$ and are then
filtered forward. Beliefs need not then be updated at each step in the
(imagined) search tree
\cite{Wang:2005:aaa,Ross:2008:aaa,Asmuth:2011:aaa}. Thus, if $T$ is the
search horizon and $K$ is the number of simulations, then {\sc bamcp}
(with root sampling) requires $O(K)$ samples from the posterior and one
belief update, instead of $O(TK)$ samples with many belief updates. For
these reasons, we chose {\sc bamcp} for our forward-search planning
algorithm in this paper.  For completeness, the {\sc bamcp} algorithm is
specified in the supplementary material; refer to
\cite{Guez:2012:aaa,Guez:2013:aaa} for more details.

\section{Statistical models of {\sc mdp}s}

There is a huge range of possible models for complex 
domains. Understanding when and how they apply is a whole
subject in its own right. Here, we adopt a strategy that has been very
successful in other areas of statistical modeling, namely using a
Bayesian non-parametric model \cite{Orbanz:2010:aaa}.  This permits
complexity to scale as observations accumulate, while carefully
parameterizing how structure is likely to repeat.

In section~\ref{sec:crpcb}, we consider a rich, non-parametric, task
that is an extension of a contextual bandit problem. However, although
solving a wholly artificial task is revealing about the differences
between different methods of planning, it says little about performance
in real cases in which the data were (likely) not generated from the
model. Thus we first motivate this rich model as a generalization of 
a realistic exploration task.

\subsection{The mushroom exploration task}
\label{sec:mushroom}

\begin{figure}[htb]
\begin{center}
  \subfigure[]{
  \raggedleft
    \includegraphics[width=.24\textwidth]{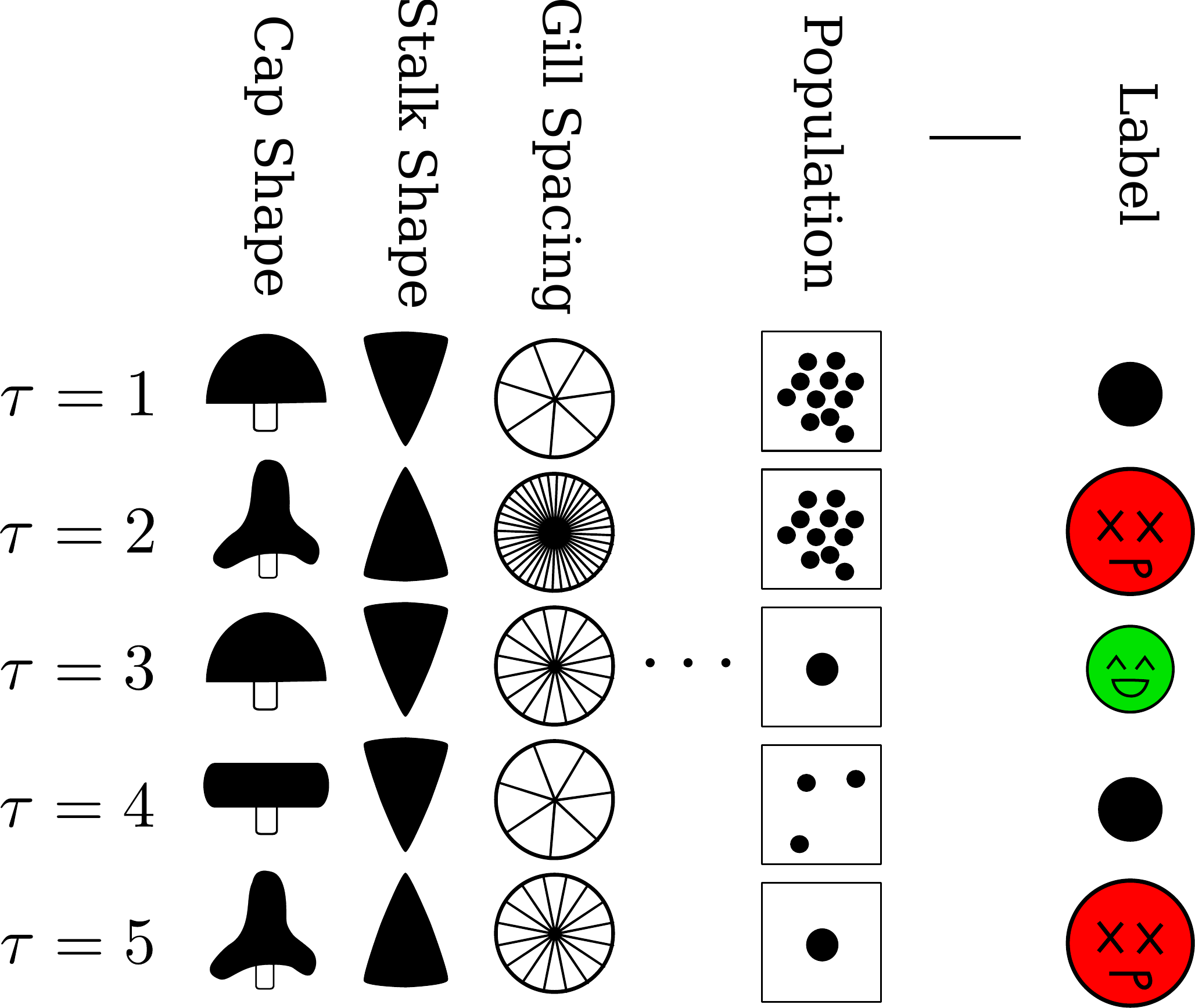}
    \hspace{0.3in}
  }
  \subfigure[]{
  \hspace{0.1in}
  \includegraphics[width=.4\textwidth]{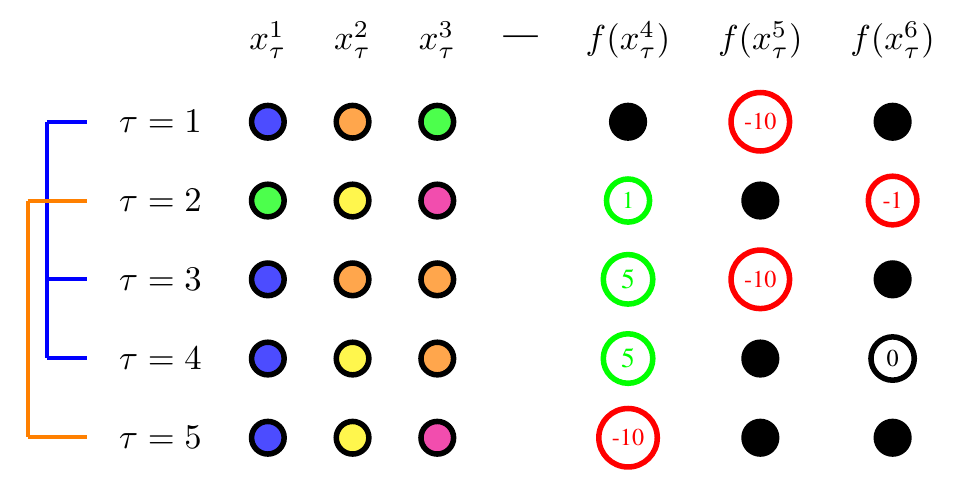}
  }  
\end{center}
\caption{\footnotesize{(a) Illustration of the mushroom exploration domain, the
rows represent the sequence of mushrooms, with the value of different
attributes being displayed on each column. The agent may eat the mushroom to
obtain a reward/cost (green/red circle) or choose to ignore it (black
circle). (b) Example of a sample sequence of tasks from Section~\ref{sec:crpcb}
with $C\!\!=\!3$, $A\!=\!3$.  The left columns indicate the context variables (each color
represent a possible value), and the right columns indicate the rewards that
have been observed in each task (black when arm has not been pulled). On the
left, the true clustering of the tasks (shared mixture component) is displayed.
The dictionary of context variables is composed of $D\!=\!5$ different colors in
this case.}}
\label{fig:historyExample}
\end{figure}

The Mushroom Dataset from the UCI repository~\cite{Bache:2013:aaa}
contains $8124$ instances of gilled mushrooms from $23$ different
species in the Agaricus and Lepiota family, each of which is described
by $22$ discrete attributes (e.g., color, odor, ring type) and whether
the mushroom is poisonous or edible (51.8\% of all instances are
edible).  We build an {\sc mdp} based on the data as follows, at each point
in time the agent is faced with the attributes of 
a random mushroom from the dataset, and 
has to choose whether to eat or ignore it. Ignoring a mushroom has no
consequence; eating an edible mushroom is rewarding ($r=5$); but eating
a poisonous mushroom incurs a large cost ($r=-15$). This is illustrated
in Figure~\ref{fig:historyExample}a. The agent may be provided with some
initial 'free' observations of the attributes and edibility of a set of
mushrooms.  

This problem is conventionally thought of in terms of supervised
learning. However, since the agent is allowed to ignore a mushroom, it
is actually more akin to a contextual bandit
task~\cite{Langford:2007:aaa}. However in our case, unlike most past
work on contextual bandits, early rewards are more valuable than later
ones, characterized by a discount factor $\gamma$.  This is a critical
difference from exploration objectives based on regret that are
dominated by the long-term behavior of the agent.

More formally, the mushroom {\sc mdp} consists of a sequence of mushroom
tasks parametrized by $\boldsymbol x_\tau$ where $\tau=1,2,\dots$.
Each parameter vector $\boldsymbol x_\tau$ contains $C=22$ scalar
parameters $x^{1}_\tau,\dots,x^{C}_\tau$ to generate context (the
mushroom attributes), and a single scalar parameter $x^{C+1}_\tau$ to
generate the subtask dynamic (the outcome of eating the mushroom).
Denoting $n=C+1$, we have $\boldsymbol x_\tau = (x^1_\tau,
x^2_\tau, \dots, x^n_\tau)$. The {\sc mdp} dynamics $\mP$ can be
described as follows. Let $S$ be the set of states, each of the form
$s = (\tau, x^{1}_\tau,\dots,x^{C}_\tau, o_\tau)$, where $o_\tau =
\Box$ (meaning unobserved) if the mushroom was not eaten in task
$\tau$ and $o_\tau = x^{C+1}_\tau$ otherwise. Choosing the exit action
$a_{\text{exit}}$ increments the first state component and updates the
context and observation components. Choosing the eat action
$a_{\text{eat}}$ updates $o_\tau = x^{C+1}_\tau$ and delivers a
reward/cost.  

\paragraph{A simple statistical model}
The key aspect of the mushroom task is the joint statistics over the
characteristics and danger of the mushrooms. The truth of the matter for
the UCI data is actually unclear; it is therefore a stringent test of a
planning algorithm whether it is possible to perform at all well based
on what can only be a vague, and likely inaccurate model.  To do its
best, the agent assumes a general non-parametric model that allows for
substantial underlying complexity in the true model, but adapts its
ongoing characterization as a function of the evidence in the data that
has so far been observed~\cite{Orbanz:2010:aaa}. We employ one
particularly popular non-parametric model called the Chinese Restaurant
Process \cite{Teh:2010:aaa} or Dirichlet Process mixture, which
postulates that the mushrooms come in a possibly infinite number of
mixture components.
 
The generative model of the mushroom statistics is formally described as
follows: \\[-10pt]
\begin{align*}
  \alpha &\sim \text{Gamma}(a,b),   
  z_\tau \sim \text{CRP}(\alpha), \  \forall \tau \in \mathbb{Z}^+, \\
  \boldsymbol\theta^i_k &\sim \text{Dirichlet}(\frac{\beta}{D_i}), \  \forall i \in \{1,\dots,n\}, \forall k \in \mathbb{Z}^+,  \\
  x^i_\tau &\sim \text{Categorical}(\boldsymbol\theta^i_{z_\tau}) \ \forall i \in \{1,\dots,n\}, \forall \tau \in \mathbb{Z}^+,
\end{align*}
where $\alpha$ is the concentration parameter of the CRP, the $z$
random variables are the cluster assignments. The base measure of this
Dirichlet process is assumed to be a symmetric Dirichlet prior with
hyperparameter $\beta=1$ ($D_i$ is the dimension of $\boldsymbol
\theta^i$, the number of possible observations for $x^i$), which
together with the conjugate observation model, allows for relatively
straightforward inference schemes (see Section~\ref{sec:inf} for
details). The collection $\{ \boldsymbol\theta^i_k | i \in
\{1,\dots,n\} \}$ of vectors contains the parameters corresponding to each
mixture component $k$, the task parameters $\boldsymbol x_\tau$ for a
particular $\tau$ are drawn by first choosing a mixture component
$z_\tau$ according to the CRP and then using the corresponding
$\boldsymbol\theta_{z_\tau}$ parameters to sample each component of
$\boldsymbol x_\tau$. The infinite-state, infinite-horizon {\sc mdp} is
derived from this generative process by sampling an infinite sequence
of tasks ($\tau \rightarrow \infty$) and patching them together. 

The data at time $\mD_t$ consist of all mushroom attributes and labels
observed (the sufficient statistics of the history of transitions),
including the current mushroom subtask (and any initial 'free'
examples). The posterior distribution over the dynamics $P(\mP | \mD_t)$
is then obtained straightforwardly from the posterior over all past and
future $\boldsymbol x_\tau$ (denoted $\boldsymbol x_{1:\infty}$),
$P(x_{1:\infty} | \mD_t)$, since $\mP$ is uniquely characterized by
$\boldsymbol x_{1:\infty}$.

For the mushroom data, we set $D_i=12$ for each context dimension $i$
--- the maximum number of values for any of the 22 attribute
dimensions in the data. This implies $2\cdot12^{22} \geq 10^{24}$
possible configurations of mushrooms assumed by the model. Since
$\alpha$ is not known, we set a generic hyperprior on $\alpha \sim
\text{Gamma}(.5,.5)$.

\paragraph{ Results in the mushroom task} 
We stress that the mushroom data were not really generated by the
process assumed in the previous section -- this is what makes the task
somewhat more realistic. Indeed, when the agent lacks prior data,
maximizing the return is highly challenging.  Randomly eating mushrooms
to sample the dataset is a particularly bad strategy because of the cost
asymmetry between edible and poisonous mushrooms. A natural point of
comparison is the policy of ignoring all mushrooms, which leads to a
neutral return of $0$.

\begin{figure}[htb]
\begin{center}
  \subfigure{
  {\small(a)} \includegraphics[width=.34\textwidth]{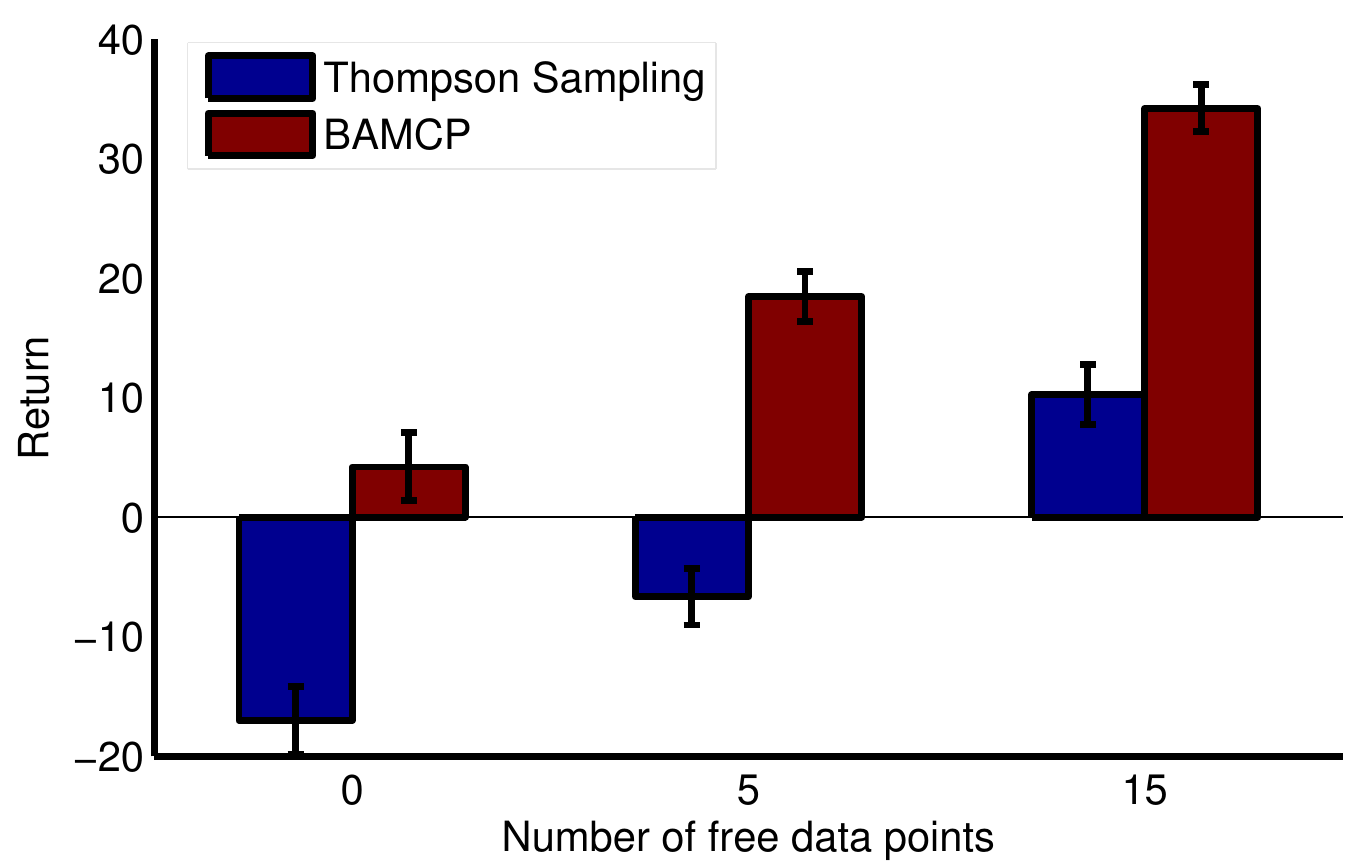}
  }
  \subfigure{
  {\small(b)} \includegraphics[width=.34\textwidth]{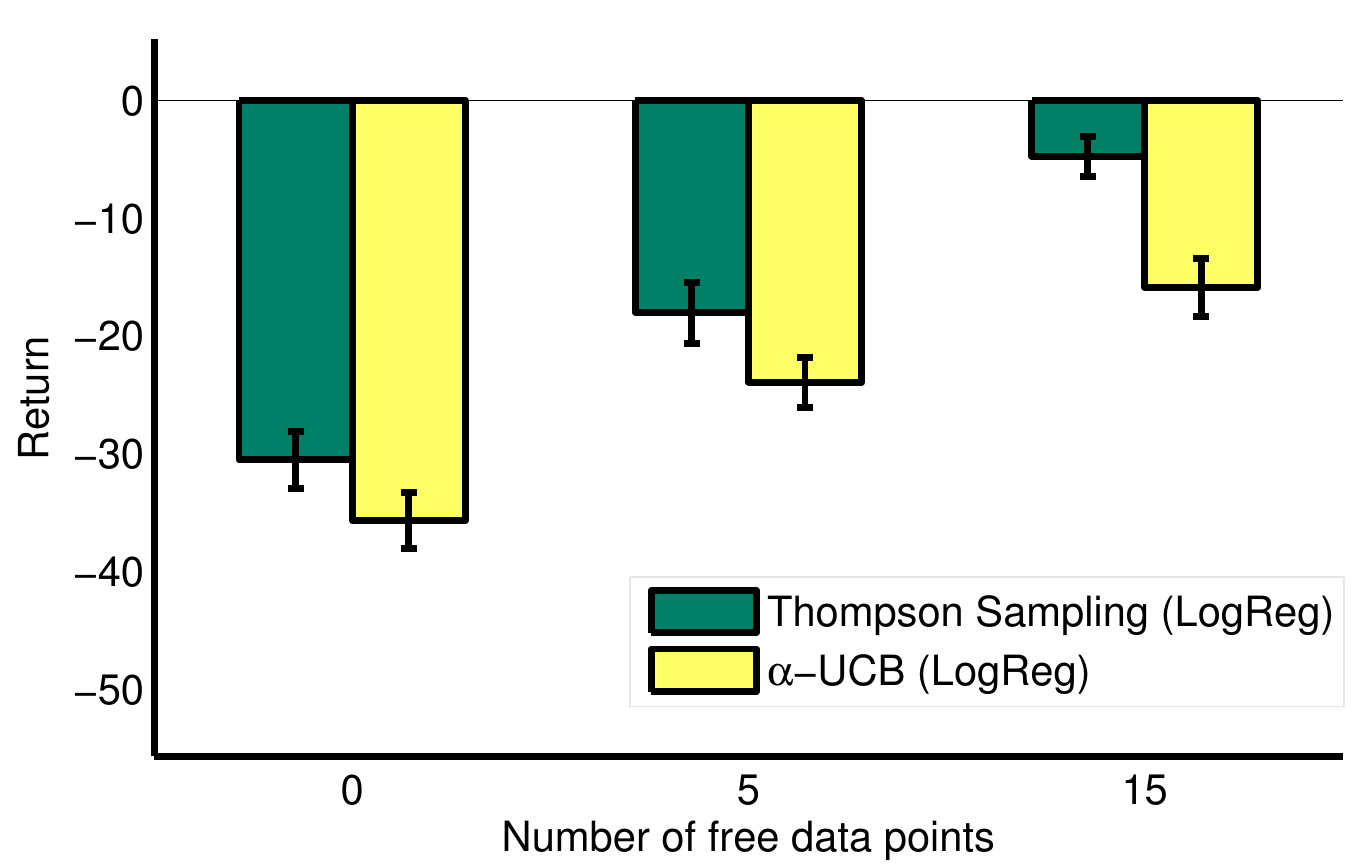}
  }
\end{center}
\vspace{-0.15in}
\caption{\footnotesize{Exploration-Exploitation results on the mushroom
dataset, after 150 steps with $\gamma\!=\!0.97$. (a) Discounted return for
{\sc bamcp} and {\sc ts} with the CRP model of
Section~\ref{sec:mushroom}, including hyperparameter inference. Either
starting from scratch from the prior (0 'free' data points), or from the
prior plus an initial random (labeled) data set of size 5 or 15. At most
75+$\{0,5,15\}$ datapoints can be observed during these 150 steps. (b)
Discounted return for {\sc ts} and $\alpha-$UCB (with
$\alpha\!=\!1$ and upper confidence approximation U0 as defined in
\cite{Li:2012:aaa}) when using the Bayesian Logistic Regression model,
same task setting as (a). Averaged over 50 runs.}}
\label{fig:mushroom}
\end{figure}

We ran the Bayes-adaptive agent ({\sc bamcp}) and {\sc ts} using this statistical
model on the mushroom task. Since the concentration parameter is
unknown, it is inferred from data, \textit{both influencing, and being
influenced by, the exploration}.  Results are reported in
Figure~\ref{fig:mushroom}a for three different numbers of 'free'
examples. A surprising result is that the Bayes-adaptive agent manages
to obtain a positive return when starting with no data, despite the
mismatch between true data and generative model. This demonstrates that
abstract prior information about structure can guide exploration
successfully. Given exactly the same statistical model, {\sc ts} fails to
match this performance; we speculate that this is due to over-optimism,
and investigate this further in Section~\ref{sec:crpcb} and 
the Supp. material
(Fig.~\ref{fig:mushexprate}). When initial data (incl. labels) is
provided for free to reduce the prior uncertainty, {\sc ts} can improve its
performance by a large margin but its return remains inferior to a
Bayes-adaptive agent in the same conditions.\footnote{We also tested the
PSRL version of {\sc ts} (Osband et al., 2013), which commits to a policy for
$\frac{1}{1-\gamma}$ steps. Performance was worse than for regular
{\sc ts}, an expected outcome, since PSRL takes more time to integrate and
react to new observations.}

For the purposes of comparison, we also considered a simpler
discriminative statistical model, namely Bayesian Logistic Regression,
which \cite{Li:2012:aaa} suggested for use on contextual bandits.
Figure~\ref{fig:mushroom} shows the results of applying {\sc ts} and
$\alpha-$UCB \cite{Li:2012:aaa} in this context. {\sc ts} does worse with the
logistic regression model than with the CRP-based model; this
demonstrates the added benefits of a prior that captures many aspects of
the data with only a few datapoints. The $\alpha-$UCB algorithm, despite
good performance in the long-run on large datasets, is too optimistic
to perform well with discounted objectives.

\subsection{Non-Parametric Contextual Bandit Sequence model}
\label{sec:crpcb}

The mushroom task can be seen as a sequence of subtasks that share
structure, but whose order the agent cannot control.  Other such domains
are adaptive medical treatments where each patient can be understood as
the subtask, handling customer interactions, or making decisions to
drill for oil at different geological locations.  In this section, we
consider a generalized version of domains with this characteristic form
of shared structure. Further, by addressing environments that were
actually drawn from the model, we study planning in the absence of model
mis-match.

The key generalization is to allow multiple arms in each subtask (rather
than a single eat/exit decision). Using the same notation as
Section~\ref{sec:mushroom}, each parameter vector $\boldsymbol x_\tau$
now contains $C$ scalar parameters $x^{1}_\tau,\dots,x^{C}_\tau$ to
generate context, and $Y$ scalar parameters
$x^{C+1}_\tau,\dots,x^{C+Y}_\tau$ to generate the actual task dynamics
(i.e., denoting $n\!=\!C\!+\!Y$, we have $\boldsymbol x_\tau =
(x^1_\tau, x^2_\tau, \dots, x^n_\tau)$). The generative model is
identical otherwise, but now the choices of the agent in any particular
task $\tau$ are to either: 1) leave the subtask for the next; or 2) pull
any of the $Y$ arms that has not been previously
pulled. 
The {\sc mdp} states are now of the form $s =
(\tau, x^{1}_\tau,\dots,x^{C}_\tau, o^{1}_\tau,\dots,
o^{Y}_\tau)$, where $o^{a}_\tau \!=\! \Box$ if arm $a$ has not been
pulled in task $\tau$ and $o^{a}_\tau \!=\! x^{C+a}_\tau$
otherwise. Figure~\ref{fig:historyExample}b shows the first part of a
draw of the generative process including an hypothetical agent
trajectory.   

The exact setting for the experiments is as portrayed in
Figure~\ref{fig:historyExample}b: with $C=3$ context cues, $A=3$ arms in
each task, and $D=5$ possible values of $x$ (i.e., the
dimension $D_i=5$ for each $\boldsymbol \theta$).  The function $f$
(that maps values of $x^a$ to rewards) is 1-1 with the domain: $\{ 5, 2,
0, -1, -10\}$. We drew {\sc mdp}s with different values of the concentration
parameter $\alpha \in \{0.1, 0.5, 1, 2, 5, 10\}$. The agent was assumed
to know the generative structure of the {\sc mdp}; but we considered both
cases when it knew the true value of $\alpha$ or just had a generic
hyperprior on $\alpha \sim \text{Gamma}(.5,.5)$, and had to learn.

This can be seen as a contextual bandit task \cite{Langford:2007:aaa}
with shared structure modeled by a CRP. The difference from the usual
definition of contextual bandit is that here, one of the arms has a
known reward of $0$ (the exit action) and that we give the option of
playing multiple arms for the same context (subtask). In addition,
unlike our algorithm, existing work on contextual bandit rarely exploits
the unsupervised learning that the context affords even when no label is
obtained. Many extensions are possible, including more complex intra-task
dynamics (we explore this avenue in the supplementary, Section~\ref{sec:crpmdp}) and more
general forms of shared structure; however we focus here on planning rather
than modeling.

\paragraph{Results on synthetic data}
\label{sec:synresults}

We investigate the behaviour and performance of Bayesian agents acting
in tasks sampled from the non-parametric model above. The reward mapping
implies that $E[f(x^a)]\!=\!-0.8\!<\!0$ for all arms $a$ and for all $\tau$,
since all values of $x^a$ are equally likely \textit{a priori}.  Thus,
again, the strategy $\pi^{\text{exit}}$ of always exiting subtasks
(without pulling any arms) is a fair comparison, with value $0$ -- a
myopic planner based on the posterior mean only should never explore an
arm, gaining this value of $0$. Any useful adaptive strategy should be
able to obtain a mean return of at least $0$.

\begin{figure}[htb]
  \begin{center}
    \subfigure{
      {\small(a)}\includegraphics[width=.35\textwidth]{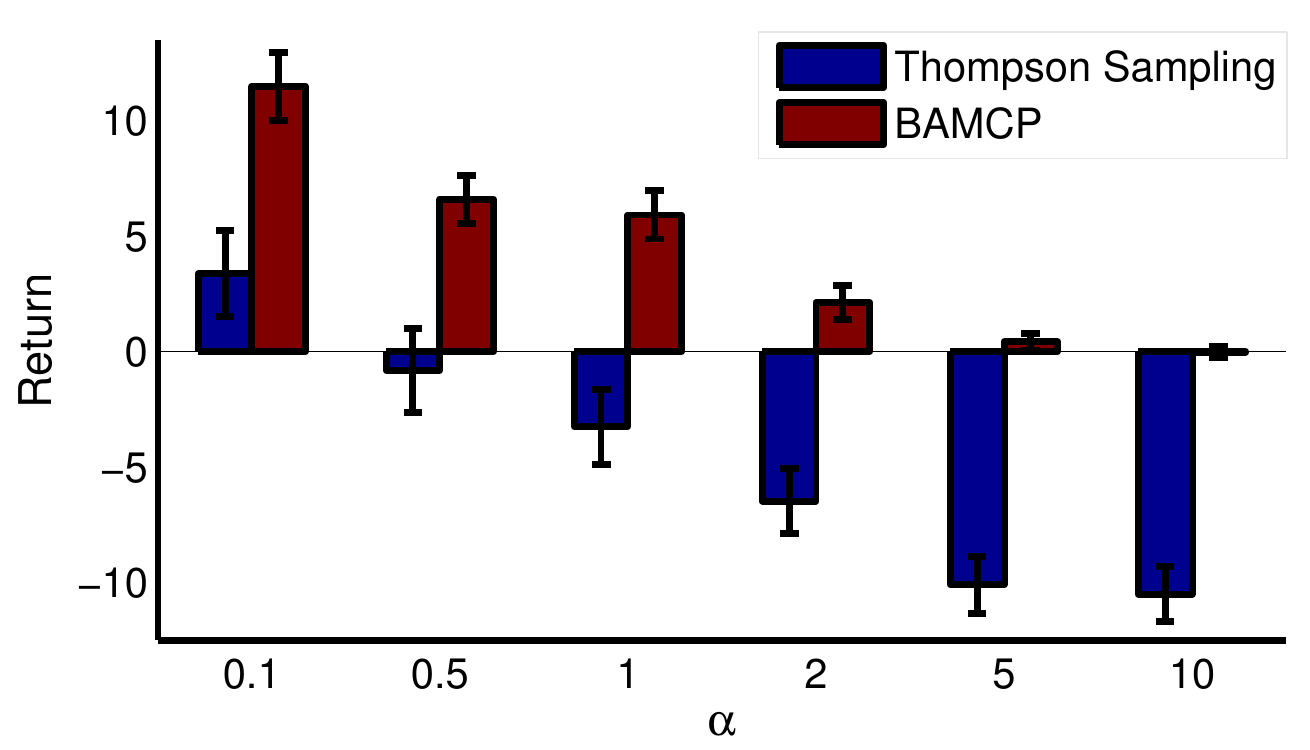}
    }
    \subfigure{
     {\small(b)}\includegraphics[width=.35\textwidth]{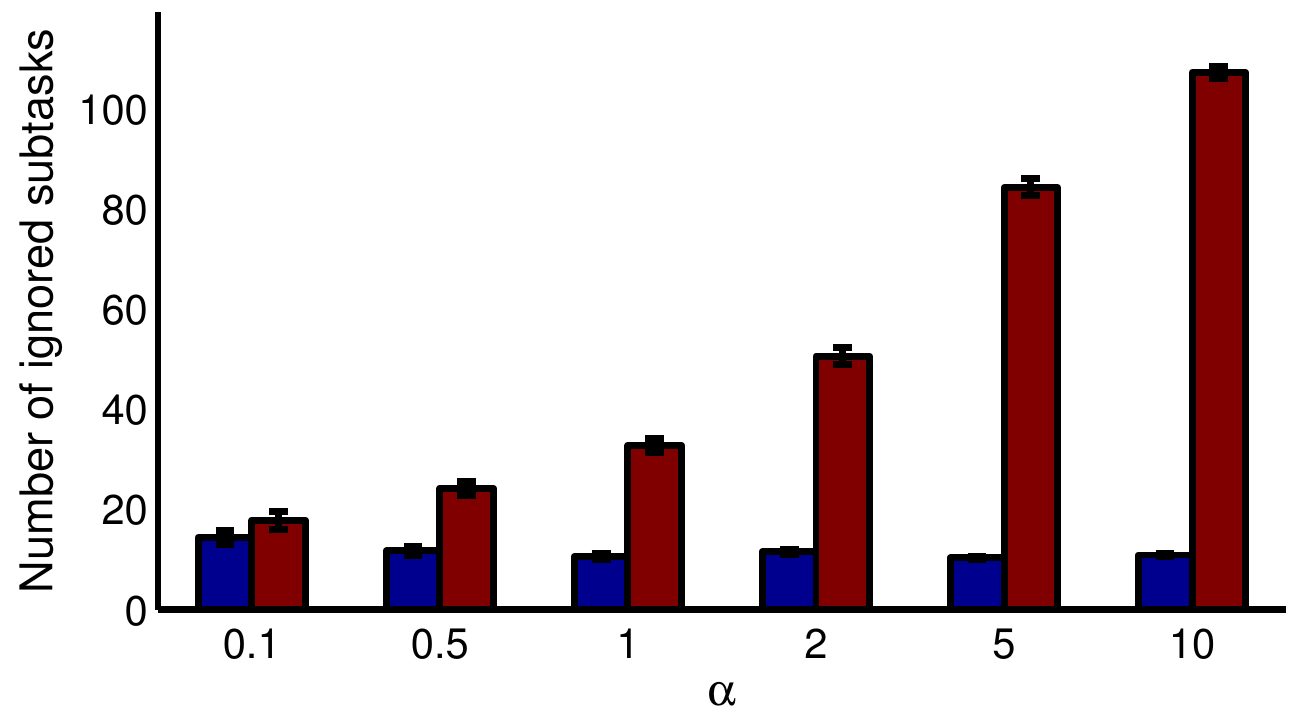}
    }
  \end{center}
  \vspace{-0.2in}
  \caption{\footnotesize{The performance of {\sc bamcp} and
      Thompson sampling on the non-parametric bandit task
      ($\gamma=0.96$) in terms of discounted return (a). The
      concentration parameter $\alpha$ is known to the algorithm and
      varies on the x-axis. In (b), the average number of subtasks
      ignored by each algorithm in the different environmental
      conditions. Each plotted value is averaged over 200 runs of 120
      steps (with 60K forward simulations per step). 
  }}
  \label{fig:crpcb}
\end{figure}

We concentrate on two metrics computed during the first $120$ steps of
the agent in the environment: the discounted sum of rewards (the formal
target for optimization; Figure~\ref{fig:crpcb}a), and the
number of times the agent decides to skip a subtask before trying any
of its arms (Figure~\ref{fig:crpcb}b). The second metric relates to
the safe exploration aspect of this task; sometimes optimism is
unwarranted because it is more likely to lead to negative outcomes,
even when taking into account the long-term consequences of the
potential information gain.

We ran {\sc bamcp} and {\sc ts} on the
task. Figure~\ref{fig:crpcb} shows the performance as a function of
$\alpha$, when this concentration parameter is known.
When the concentration parameter $\alpha$ is small, there will only be a
few different mixture components, making for an easy case with little
uncertainty about the identity of the mixture components after a few
observations, and therefore little uncertainty about the outcome of an
arm pull. In the limit of $\alpha \rightarrow 0$, only one cluster will
exist and the domain essentially degenerates to a form of multinomial
multi-armed bandit problem. As $\alpha$ grows, the identity of a given
task's cluster becomes more uncertain and aliasing grows, so safe
exploration becomes more challenging. Learning is slower in that regime
too, simply because there are more parameter values to acquire. As
$\alpha \rightarrow \infty$, every cluster will be different; this would
prevent any kind of generalization and the Bayes-optimal policy will be
to skip every subtask $\tau$ (since the \textit{a priori} expected
values of the arms in any given subtask is negative).

Figure~\ref{fig:crpcb} shows that {\sc bamcp} adapts its
exploration-exploitation strategy according to the structure in the
environment; small values of $\alpha$ justify the risk of exploring and
incurring costs but this optimism progressively disappears as $\alpha$
gets larger. This translates into positive return when generalization is
feasible, despite the marginal negative expected cost for each arm, and
a return close to $0$ when costs cannot be avoided. On the other hand,
{\sc ts} suffers from over-optimism across the board, leading to poor
discounted returns, especially when the number of mixture components is
large. Intuition for {\sc ts}'s poor performance comes from considering an
extreme case in which all or most subtasks are sampled from a different
cluster. Here, past experience provides little information about the
value of the arms for the current cluster; thus, discovering these
values (which, on average, is expensive) is not likely to help in the
future. However, {\sc ts} samples a single configuration of the arm, mostly
informed by the prior in this situation, which likely results in at
least \textit{one of the arms} as having a positive outcome (for the
prior, we repeat 3 times a draw having $\frac{2}{5}$ probability of
success, so $p=0.784$). {\sc ts} then, incorrectly, picks this putatively
positive arm rather than exiting. Other myopic sample-based exploration
strategies, such as Bayesian DP~\cite{Strens:2000:aaa} or
BOSS~\cite{Asmuth:2009:aaa}, would suffer from similar forms of
unwarranted optimism --- since they also rely on sampling one or more
posterior samples according to which they act greedily (see Examples
3-4). 

\begin{figure}[ht]
\begin{center}
  \subfigure{
  {\small(a)}\includegraphics[width=.34\textwidth]{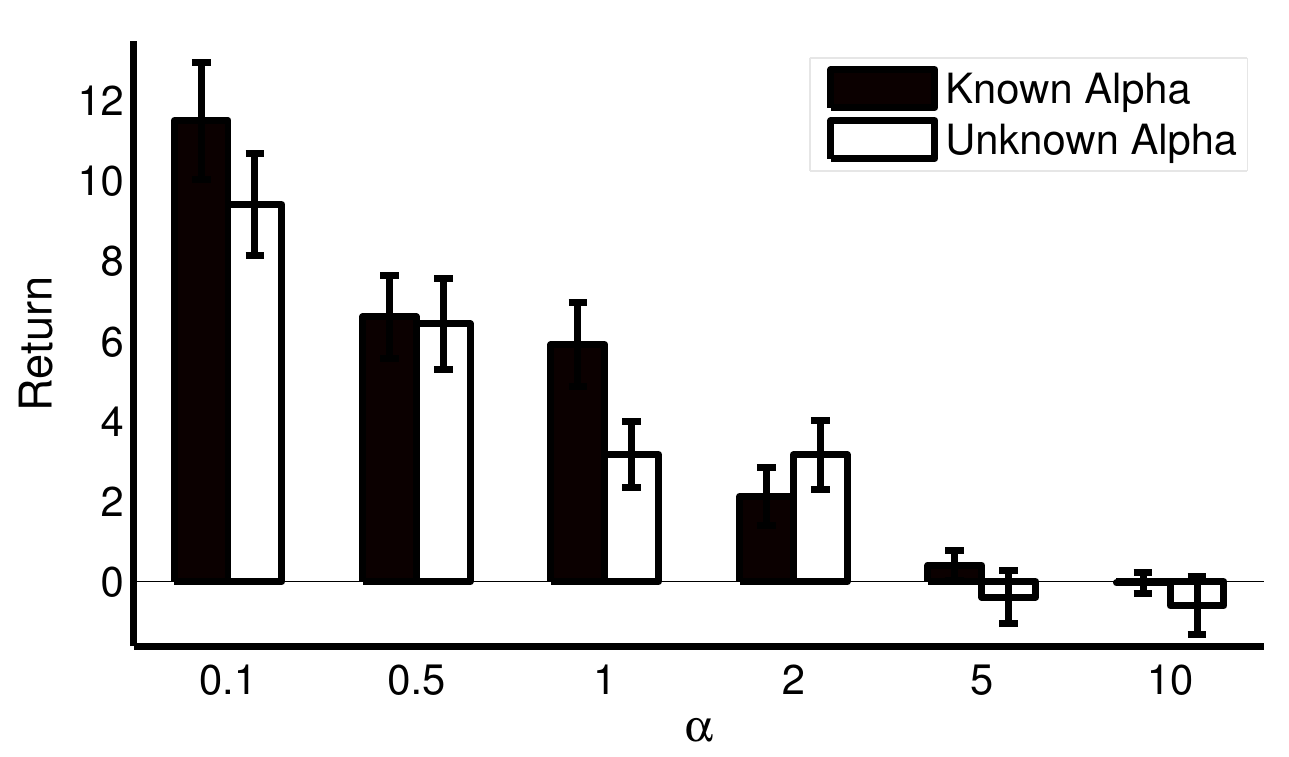}
  }
  \subfigure{
    {\small(b)}\includegraphics[width=.34\textwidth]{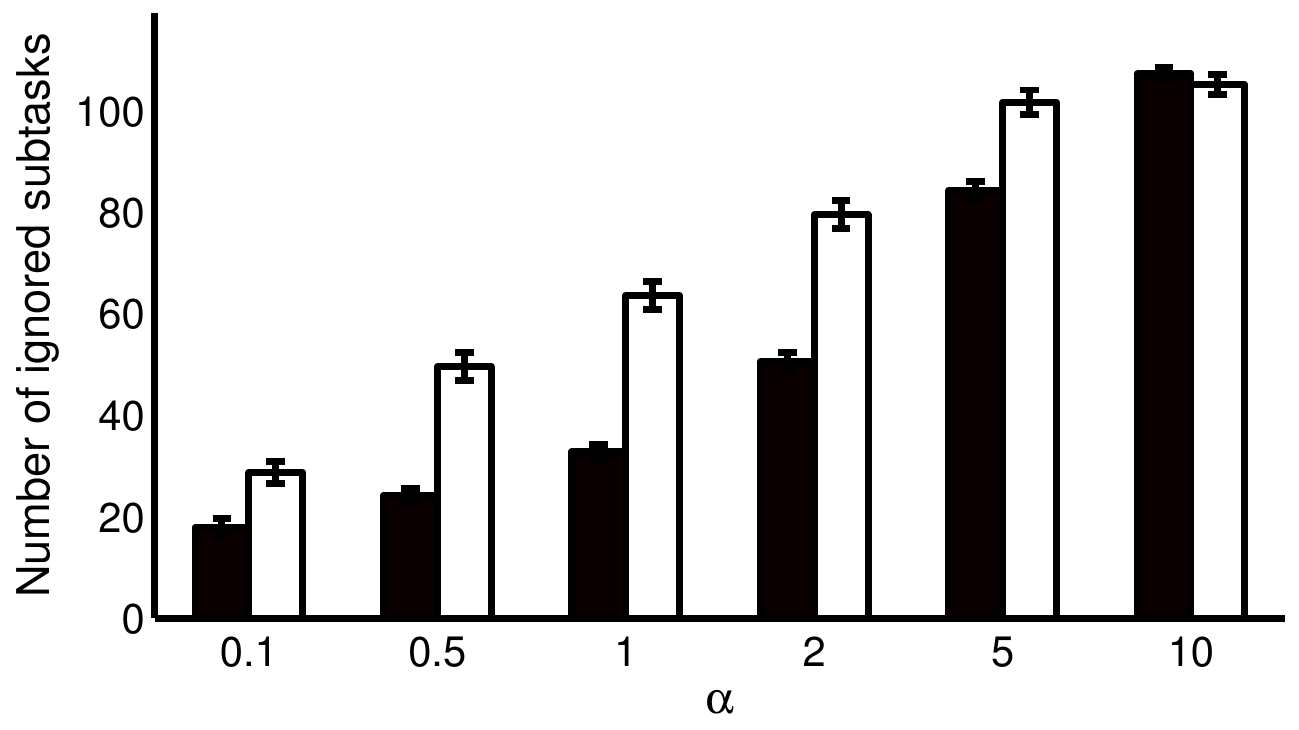}
  }
\end{center}
\vspace{-0.2in}
\caption{\footnotesize{Performance of {\sc bamcp} for various values of the
    concentration parameter $\alpha$, with and without hyperparameter
    inference. The learned Bayes-adaptive policy avoids more subtasks
    (b) but manages to maintain a similar level of performance (a)
    despite the uncertainty over $\alpha$. Averaged over 200 runs with
    60K simulations for each time-step. The discount parameter is
    $\gamma=0.96$.}} 
\label{fig:known_vs_inf}
\end{figure}

A yet more challenging scenario arises when $\alpha$ is not known to the
agent. A Bayesian agent starts with a uninformative (hyper-)prior on
$\alpha$ in order to infer the value from data, it also takes into
account during planning how its belief about this hyperparameter changes
over time. In Figure~\ref{fig:known_vs_inf}, we observe that the
Bayes-adaptive agent is more conservative and explores more safely when
$\alpha$ is unknown. As expected, this results in lower returns
(compared to when $\alpha$ is known). However, robustness to increased
uncertainty is shown by the modest difference.

In the suppl.\ material (Fig.~\ref{fig:gamcomp}), we also show that
{\sc bamcp} is sensitive to the discount factor, highlighting the dependence
of the exploration-exploitation strategy to the horizon.

\section{Related Work}

Many researchers have considered powerful statistical models in the
context of sequential decision-making~\cite{Wingate:2011:aaa,Lazaric:2010:aaa}, including in exploration-exploitation
settings~\cite{Doshi-Velez:2010:aaa,Asmuth:2009:aaa}. Non-parametric
models have been considered in the context of control
before~\cite{Doshi-Velez:2009:aaa,Asmuth:2009:aaa} but with an
emphasis on modeling the data rather than planning. In \cite{Ross:2008:aaa},
the authors consider factored {\sc mdp}s whose transitions are modeled using
Bayesian Networks. They demonstrate the advantages of having an appropriate
prior to capture the existing structure in the true dynamics, at least in a
case in which the problems of safe exploration do not arise. 
For planning, they propose an online
Monte-Carlo algorithm with an approximate sampling scheme, however the
forward-search is conducted with a depth of 2 and a small branching
factor, presumably limiting the benefits of Bayes-adaptivity.    

\cite{Moldovan:2012:aaa} consider a particular form of safe
exploration to deal with non-ergodic {\sc mdp}s, but they do not address
discounted objectives or structured models.

\cite{Guez:2012:aaa} consider an infinite {\sc mdp}, combining
Bayes-adaptive planning with approximate inference over possible {\sc
mdp}s. However the class of models is quite specific to the particular
domain they consider. In \cite{Doshi-Velez:2009:aaa}, an hierarchical
Dirichlet Process is used to allow for an unbounded number of states in
a {\sc pomdp} and infer the size of the state space from data, this is
referred as the i{\sc pomdp} model. This model is used in a online
forward-search planning scheme, albeit of rather limited depth and
tested on modestly-sized problems.

In \cite{Deisenroth:2011:aaa}, Gaussian Processes (GPs) are employed to
infer models of the dynamics from limited data, with excellent empirical
performance. However, the uncertainty that the GP captures was not
explicitly used for exploration-exploitation-sensitive planning. This is
addressed in \cite{Jung:2010:aaa}, but with heuristic planning based on
uncertainty reduction.

More generally, our task is reminiscent of the case of active
classification~\cite{Balcan:2009:aaa}. But while active learning
ultimately aims to find an accurate classifier on a labeling budget,
we are concerned with a completely different metric, namely discounted
return. In particular, a perfectly fine solution in our setting might be
to avoid labeling a large part of the input space.

\section{Discussion}

Model-based Bayesian RL has often been viewed as attractive yet
hopeless, particularly in high-dimensional and noisy domains. It had
previously been shown that {\sc bamcp}, an efficient combination of
extensions to Monte-Carlo tree search for Bayes-adaptive planning, was
computationally viable, and yet very powerful. 
We showed that alternative,
over-optimistic, myopic planning methods such as Thompson Sampling can run into
severe problems that {\sc bamcp} avoids through explicit lookahead computations.

In an attempt to scale Bayes-adaptive planning to real domains, we
proposed a contextual-bandit benchmark domain derived from the UCI
mushroom dataset and an associated Bayesian non-parametric model for
it. In this challenging exploration-exploitation domain, we
demonstrated the feasibility and advantages of using a Bayes-Adaptive,
or fully Bayesian, agent.  

There are various ways to improve planning. Along with generic ideas
such as adaptive adjustments of the roll-out policy which exerts a
strong influence over the performance of {\sc bamcp}, it would be
interesting to think about more radical departures, such as function
approximation within the search tree based on histories and possible
future histories.

A remaining open problem is to understand which classes of domains will
truly benefit from the computations of Bayes-Adaptive planning (such as the
ones explored in this paper), and which will be served just right with a
simpler exploration-exploitation approach. Indeed, one can come up with
examples where additional computation barely matters, in that the gains are
vanishingly small (e.g., the Gittins indices for a multi-armed bandit problem
with $\gamma\approx 1$ are hard to obtain, but many myopic policies
would do well in that scenario). 

We have focused on planning; this means that the challenge now opened up by the
success of {\sc bamcp} is modeling. The non-parametric model of shared
structure amongst sub-tasks is readily generalizable to many domains, including
ones in which the equivalent of the arms are themselves {\sc mdp}s (see
Section~\ref{sec:crpmdp}). A more radical extension would be to something
closer to an Indian buffet process~\cite{Griffiths:2011:aaa}, in which the
whole collection of subtasks also share a measure of structure; this should
lead to solutions with collaboration among a set of expert solvers.

\begin{footnotesize}
	\bibliography{paper}
\end{footnotesize}
\bibliographystyle{aaai}

\clearpage
\setcounter{figure}{0}
\setcounter{section}{0}
\makeatletter 
\renewcommand{\thefigure}{S\@arabic\c@figure}
\renewcommand{\thesection}{S\arabic{section}}   

{\Large \textbf{Supplementary Material}}

\section{Bayes-Adaptive Planning}
\label{sec:bap}
We describe the generic Bayesian formulation of optimal decision-making in an unknown Markov Decision Process ({\sc mdp}), we refer the reader to \cite{Martin:1967:aaa} and \cite{Duff:2002:aaa} for additional details. An {\sc mdp} is described as a 5-tuple $M = \langle S, A,\mathcal{P},\mathcal{R},\gamma \rangle$, where $S$ is the set of states, $A$ is the set of actions, $\mathcal{P} : S \times A \times S \rightarrow \mathbb{R}$ is the state transition probability kernel, $\mathcal{R} : S \times A \rightarrow \mathbb{R}$ is a bounded reward function, and $\gamma$ is the discount factor~\cite{Szepesvari:2010:aaa}.  When all the components of the {\sc mdp} tuple are known, standard {\sc mdp} planning algorithms can be used to estimate the optimal value function and policy off-line. In general, the dynamics are unknown, and we assume that $\mathcal{P}$ is a latent variable distributed according to a distribution $P(\mathcal{P})$.  After observing a history of actions and states $\mD_t=s_1a_1s_2 a_2 \dots a_{t-1} s_t$ from the {\sc mdp}, the posterior belief on $\mathcal{P}$ is updated using Bayes' rule $P(\mathcal{P}| \mD_{t}) \propto P(\mD_{t} | \mathcal{P}) P(\mathcal{P})$. The uncertainty about the dynamics of the model can be transformed into uncertainty about the current state inside an augmented state space $S^+ = \mathbfcal{D}$, where $\mathbfcal{D}$ is the set of possible histories. The dynamics associated with this augmented state space are described by 
\begin{align}
\mathcal{P}^+(\mD, a, \mD as') &= 
  \int_{\mathcal{P}} \mathcal{P}(s,a,s') P(\mathcal{P} | \mD) \, \text{d} \mathcal{P},\\
\mathcal{R}^+(\mD ,a) &= R(s,a),
\label{eq:BAMDP}
\end{align}
where $s$ is the suffix state of $\mD$. Together, the 5-tuple $M^+ = \langle S^+, A,\mathcal{P}^+,\mathcal{R}^+,\gamma \rangle$ forms the Bayes-Adaptive {\sc mdp} ({\sc bamdp}) for the {\sc mdp} problem $M$. Since the dynamics of the {\sc bamdp} are known, it can in principle be solved to obtain the optimal value function associated with each action: 
\begin{equation}
  Q^*(\mD_t, a) = \max_\pi \mathbb{E}_\pi \left[ \sum_{t'=t}^{\infty} \gamma^{t'-t} r_{t'} | a_t = a\right],
\label{eq:optQ}
\end{equation} 
where $\pi$ is a policy over the augmented state space, from which the optimal action for each belief-state can be readily derived. Optimal actions in the {\sc bamdp} are executed greedily in the real {\sc mdp} $M$ and constitute the best course of action for a Bayesian agent with respect to its prior belief over $\mathcal{P}$. It is obvious that the expected performance of the {\sc bamdp} policy in the {\sc mdp} $M$ is bounded above by that of the optimal policy obtained with a fully-observable model, with equality occurring, for example, in the degenerate case in which the prior only has support on the true model.  

\section{Examples - Cont.}

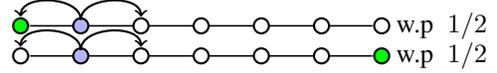
\begin{figure}[htb]
\begin{center}
\begin{tikzpicture}[scale=0.8]
    \node (X) at (0,0) {};
    \node (Y) at (6,0) {};
    \draw [semithick] (X) -- (Y);
    \node[draw,circle,thick,inner sep=0pt,minimum size=0.2cm,fill=green] at (0,0) {};
    \node[draw,circle,thick,inner sep=0pt,minimum size=0.2cm,fill=blue!30] at (1,0) {};
    \node[draw,circle,thick,inner sep=0pt,minimum size=0.2cm,fill=white!30] at (2,0) {};
    \node[draw,circle,thick,inner sep=0pt,minimum size=0.2cm,fill=white!30] at (3,0) {};
    \node[draw,circle,thick,inner sep=0pt,minimum size=0.2cm,fill=white!30] at (4,0) {};
    \node[draw,circle,thick,inner sep=0pt,minimum size=0.2cm,fill=white!30] at (5,0) {};
    \node[draw,circle,thick,inner sep=0pt,minimum size=0.2cm,fill=white!30] at (6,0) {};
    \node[draw=none] at (7,0) {w.p $\;1/2$};
    \coordinate (A) at (0,0.15);
    \coordinate (B) at (1,0.1);
    \coordinate (C) at (2,0.15);
    \draw [thick,black,->]   (B) to[out=90,in=90] (A);
    \draw [thick,black,->]   (B) to[out=90,in=90] (C);
    
    \node (X) at (0,-0.5) {};
    \node (Y) at (6,-0.5) {};
    \draw [semithick] (X) -- (Y);
    \node[draw,circle,thick,inner sep=0pt,minimum size=0.2cm,fill=white!30] at (0,-0.5) {};
    \node[draw,circle,thick,inner sep=0pt,minimum size=0.2cm,fill=blue!30] at (1,-0.5) {};
    \node[draw,circle,thick,inner sep=0pt,minimum size=0.2cm,fill=white!30] at (2,-0.5) {};
    \node[draw,circle,thick,inner sep=0pt,minimum size=0.2cm,fill=white!30] at (3,-0.5) {};
    \node[draw,circle,thick,inner sep=0pt,minimum size=0.2cm,fill=white!30] at (4,-0.5) {};
    \node[draw,circle,thick,inner sep=0pt,minimum size=0.2cm,fill=white!30] at (5,-0.5) {};
    \node[draw,circle,thick,inner sep=0pt,minimum size=0.2cm,fill=green] at (6,-0.5) {};
    \node[draw=none] at (7,-0.5) {w.p $\;1/2$};
    \coordinate (D) at (0,-0.35);
    \coordinate (E) at (1,-0.4);
    \coordinate (F) at (2,-0.35);
    \draw [thick,black,->]   (E) to[out=90,in=90] (D);
    \draw [thick,black,->]   (E) to[out=90,in=90] (F);
\end{tikzpicture}
\vspace{-0.1in}
\caption{\footnotesize{Illustration of Example 2.}}
\label{fig:ex2}
\end{center}
\end{figure}

\paragraph{Example 2} \textit{Consider a slight modification of Example 1
where the start state is the second state on the chain, so that with
probability $\frac{1}{2}$ the agent is only 1 step away from the reward
source (See Figure~\ref{fig:ex2}). Clearly, the Bayes-optimal solution is to go left towards
the nearest end first. The value of the policy at the start state is
$V^* = \frac{1}{2} (\gamma + \gamma^{2x+1})$. On the other hand,
an algorithm that commits to a particular policy based on
a posterior sample will aim for the left or right end of the chain
equally often (since they are equally likely). The resulting
value of such a strategy is $V=\frac{1}{4}(\gamma + \gamma^{2x-2} (1 + \gamma^3) + \gamma^{4x-1})$,
which gives $V= \frac{1}{2} V^*$ as $x \rightarrow \infty$.}

\vspace{-0.1in}
\paragraph{Example 3} 

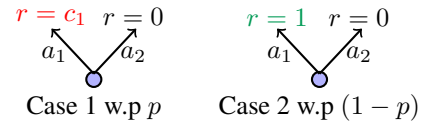
\begin{figure}[htb]
\begin{center}
\begin{tikzpicture}[scale=0.75]
    \node[draw=none] at (2,-0.6) {Case 1 w.p $p$};
    \coordinate (A) at (1.25,0.8);
    \coordinate (B) at (2,0);
    \coordinate (C) at (2.75,0.8);
    \node[draw,circle,thick,inner sep=0pt,minimum size=0.2cm,fill=blue!30] at (2,-0.1) {};
    \draw [thick,black,->]   (B) to (A);
    \draw [thick,black,->]   (B) to (C);
    \node[draw=none] at (1.3,0.35) {$a_1$};
    \node[draw=none] at (2.7,0.35) {$a_2$};
    \node[draw=none] at (1.25,1.0) {\red{$r=c_1$}};
    \node[draw=none] at (2.7,1.025) {$r=0$};
   
    \node[draw=none] at (6,-0.6) {Case 2 w.p $(1-p)$};
    \coordinate (A) at (5.25,0.8);
    \coordinate (B) at (6,0);
    \coordinate (C) at (6.75,0.8);
    \node[draw,circle,thick,inner sep=0pt,minimum size=0.2cm,fill=blue!30] at (6,-0.1) {};
    \draw [thick,black,->]   (B) to (A);
    \draw [thick,black,->]   (B) to (C);
    \node[draw=none] at (5.3,0.35) {$a_1$};
    \node[draw=none] at (6.7,0.35) {$a_2$};
    \node[draw=none] at (5.25,1.0) {\color{ForestGreen}{$r=1$}};
    \node[draw=none] at (6.7,1.0) {$r=0$};
\end{tikzpicture}
\vspace{-0.1in}
\caption{\footnotesize{The two possible 
payoff structures of Example 3.}}
\label{fig:ex3}
\end{center}
\end{figure}

\textit{Consider a single-step decision
between two actions, $a_1$ and $a_2$, with uncertainty in the payoff as
follows. With probability $p$ (case 1), action $a_1$ leads to reward
$c_1<0$ and $a_2$ leads to reward $0$. With probability $1-p$ (case 2), action
$a_1$ leads to reward $1$ and $a_2$ leads to negative reward $0$. This
is illustrated in Figure~\ref{fig:ex3}.}

\vspace{-0.05in}

\vspace{-0.1in}
\textit{Conventional {\sc ts} in this example  involves sampling one of the transitions
according to the prior and taking the corresponding optimal action. This results
in the following expected reward: $V_{TS} = E[r] = p ( p \cdot 0 + (1-p) \cdot 0) + (1-p) (p \cdot c_1 + (1-p) \cdot 1) = (1-p) (p \cdot c_1 + (1-p))$.
If $c_1$ is arbitrarily large and negative, $V_{TS}$ can be made arbitrarily bad.
The Bayes-optimal policy integrates over the possible outcomes,
therefore it performs at least as well as always choosing action $a_2$
with an expected reward of $0$. This implies $V^* \geq 0$.
The BOSS algorithm~\cite{Asmuth:2009:aaa} constructs an 
optimistic {\sc mdp} based on $K$ posterior samples, so that the best action
across all $K$ samples is taken. In this example, it is enough for
a single sample of case 2 to be present in these $K$ samples to decide
to take action $a_1$ (since $r_2 > r_1$), resulting in the following
value for BOSS (denoting $X$ to be the number of samples in the 
set of $K$ samples of case 2):
$V_{BOSS}^{ex3} = P(X \geq 1) (p \cdot c_1 + (1-p)) 
           = (1-p^K) (p \cdot c_1 + (1-p)) := z(K)$,
which is a decreasing function of $K$ (since $c_1<0$), showing 
the cost of this added optimism.}

Of course, we usually think of BOSS as being applied to {\sc mdp}s with
sequential decisions, but one can readily transform Example 3
in an {\sc mdp} by putting these 1-step decisions one after the other.
We provide details of the construction in Example 4.

\vspace{-0.1in}
\paragraph{Example 4}

\textit{Consider linking together different instances of Example 3. The agent starts in
$s_0$, and chooses between $a_1$ and $a_2$ with payoff described in Example 3.
After executing either action, the agent makes a transition to state $s_1$,
where the process repeats until state $s_n$, which itself transits back to
$s_0$. The outcome of $a_1$ and $a_2$ (determined by whether $s_i$ is of case 1 or 2) is
independent across states.}

\textit{BOSS has a parameter $B$ that decides the number of steps between posterior
resampling operations. However, $B$ has no effect in this example for the first
$n$ steps. To compute the value of BOSS's policy from the initial belief state,
leveraging the independence assumption, we can employ the value analysis of
Example 3 for the first visit of every state.  After every state gets visited
once, the transition in some states (where action $a_1$ was chosen) will be
uniquely known and BOSS can perfectly exploit the {\sc mdp} in these states. For
simplicity, we bound the value of the policy by assuming perfect knowledge of
the {\sc mdp} after the first $n$ states:
$V_{BOSS}^{ex4} < \sum_{t=0}^n \gamma^t z(K) + \sum_{t=n+1}^\infty \gamma^t (1-p) 
  = z(K) \frac{1-\gamma^n}{1-\gamma} +  \frac{\gamma^{n+1}(1-p)}{1-\gamma}$,
where $z(K) = (1-p^K) (p \cdot c_1 + (1-p))$ (from Example 3). We can choose
$n$ large to make the second term arbitrarily small, whereas the first term
again depends on $c_1<0$, which we can make arbitrarily bad.
Again, the value of the Bayes-optimal policy in this example can easily be
lower-bounded as $V^* \geq 0$ since the Bayes-optimal policy can at least choose
action $a_2$ at all times (no exploration).}

We stress that PSRL and BOSS do enjoy strong theoretical guarantees for
a different objective, namely expected regret; our goal of Bayes
adaptivity is a more severe objective because discounting exerts
pressure to perform well within a relatively shorter time horizon.

\begin{figure}[htpb]
  \begin{center}
    \includegraphics[width=.4\textwidth]{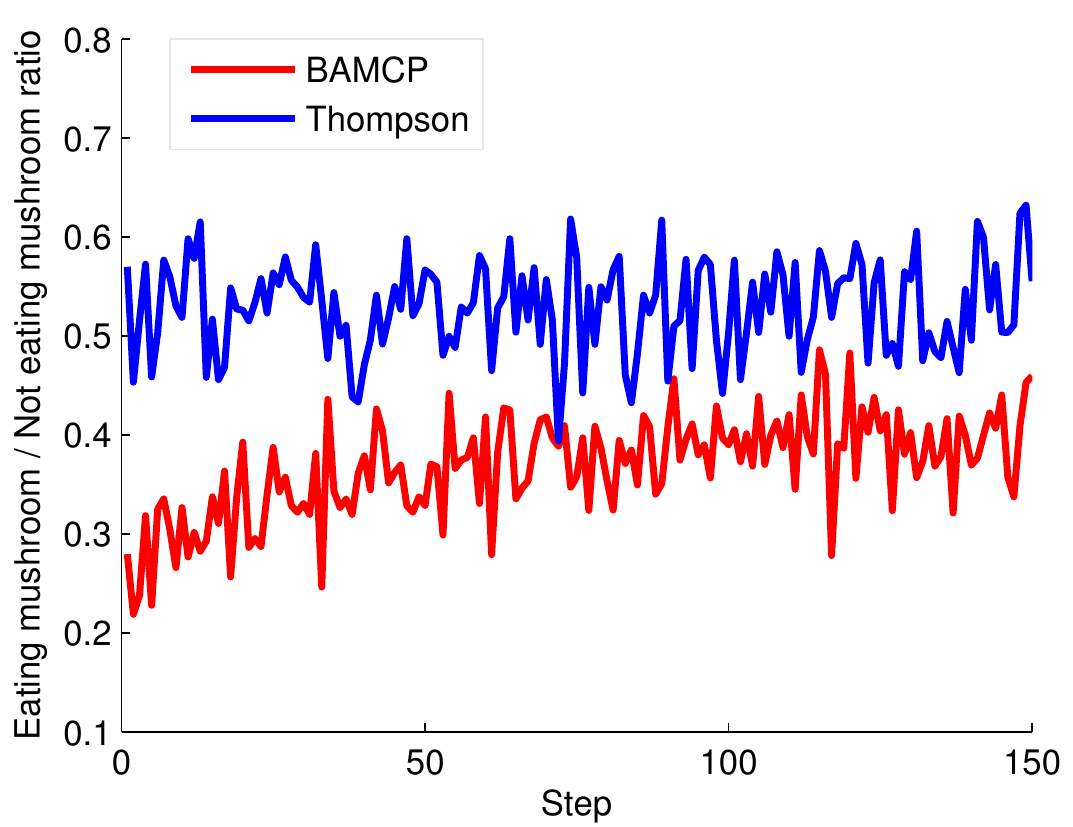}
  \end{center}
\caption{\footnotesize{Average rate of exploitation of mushrooms over
time for {\sc bamcp} and {\sc ts} for the domain described in
Section~\ref{sec:mushroom}, when starting with no labelled observations (0
free data condition). For a given step, the reported value is the
fraction of time the agent chose to eat the current mushroom (when it
had the option to), rather than to skip it.}}
\label{fig:mushexprate}
\end{figure}

\begin{figure}[htpb]
  \begin{center}
    \includegraphics[width=.35\textwidth]{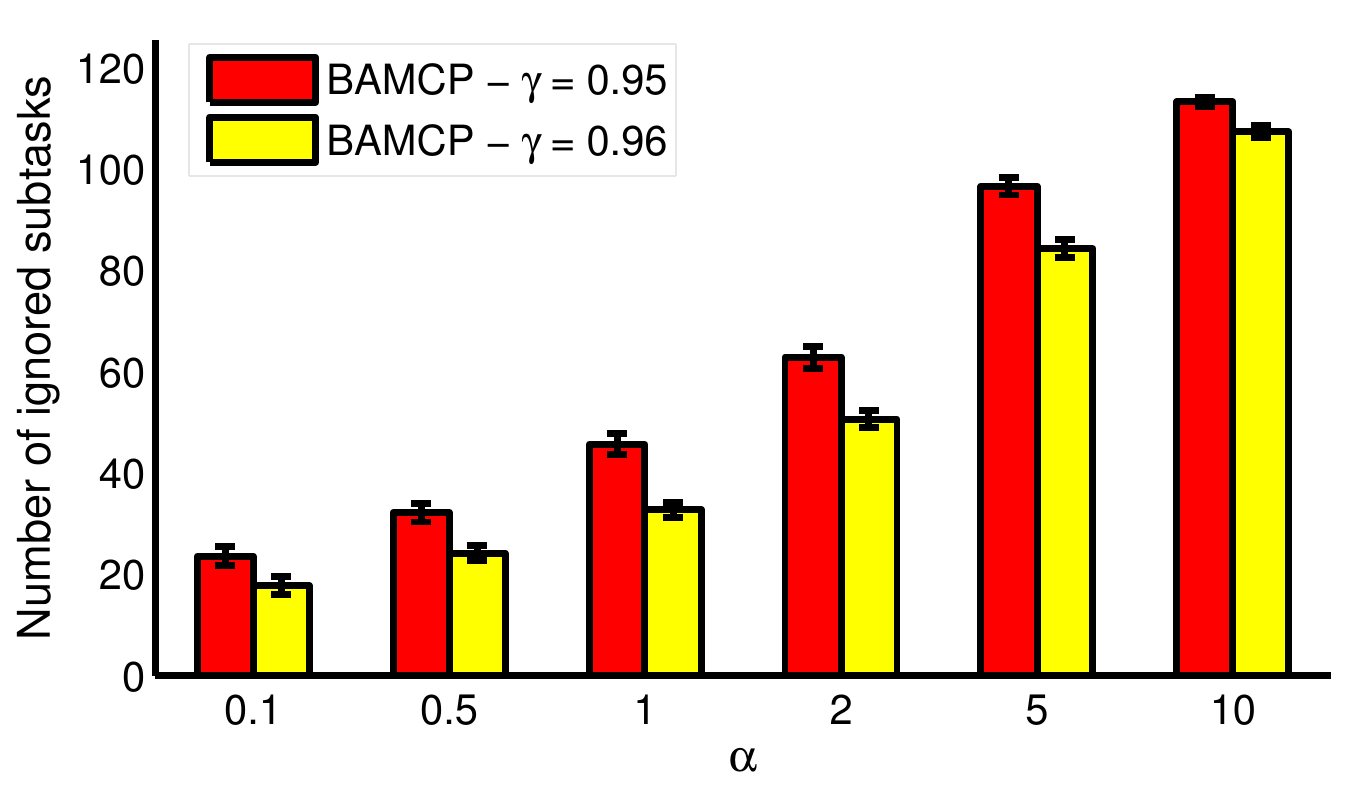}
  \end{center}
\caption{\footnotesize{In the domain described in
Section~\ref{sec:synresults}, the average number of ignored subtasks
over 120 steps when using {\sc bamcp} with either $\gamma=0.95$ or
$\gamma=0.96$ (Known $\alpha$ scenario). We see that a larger $\gamma$
induces more exploration and risk-taking (fewer subtasks are ignored),
showing the sensitivity of Bayes-Adaptive planning to the horizon.}}
\label{fig:gamcomp}
\end{figure}

\vspace{-0.1in}
\section{Inference details for the CRP-based contextual-bandit sequence model}
\label{sec:inf}
\begin{itemize}
  \item We use a Rao-Blackwellized Gibbs sampler, as described for example in \cite{Sudderth:2006:aaa}. We use the auxiliary variable trick from \cite{Escobar:1995:aaa} for tractable inference of the concentration parameter $\alpha$.
  \item A couple of Gibbs sweeps are performed between every {\sc bamcp} simulation
    but this thinning is not necessary for convergence. For Thompson Sampling, we burn in the chain with $500$ Gibbs sweep before selecting the sample used for a particular step. 
  \item Given a setting of the cluster assignments and cluster parameters for the observed subtasks (obtained from the Gibbs sampler), the future subtasks $\boldsymbol x_{t:\infty}$ are sampled by running the generative model forward conditioned on the inferred variables.
  \item In order to improve the planning speed, posterior samples can be memoized at the root of the search tree; the
  simulations pick from a pool of previously generated samples. The pool gets slowly regenerated.  
  \end{itemize}

\begin{figure}[htbp]
\removelatexerror
\begin{algorithm2e}[H]
\caption{{\sc bamcp}}
\label{alg:bmcp}
\DontPrintSemicolon
\KwSty{procedure} \FuncSty{Search(} $\mD$ \FuncSty{)} \; 
\Indp \Repeat{\FuncSty{Timeout()}}{
$\mathcal{P} \sim P(\mathcal{P} | \mD)$ \;
\FuncSty{Simulate($\mD, \mathcal{P}, 0$)}\;}
\Return $\displaystyle\argmax_a \; Q( \mD, a)$\;
\Indm
\KwSty{end procedure}\;
$ $\; 
\KwSty{procedure} \FuncSty{Rollout(}\!
                  $\mD, \mathcal{P}, d$ \FuncSty{)} \;
\Indp \If{$\gamma^{d} Rmax < \epsilon$}{
	\Return $0$\;
}
$s \leftarrow $ last state in $\mD$\;
$a \sim \pi_{ro}(\mD, \cdot)$\;
$s' \sim \mathcal{P}(s,a,\cdot)$\;
$r \leftarrow \mathcal{R}(s,a)$\;
\Return $r\!+\!\gamma$\FuncSty{Rollout(} $\mD\!as', \mathcal{P},d\!+\!1$\FuncSty{)} \;
\Indm
\KwSty{end procedure}\;
$ $\; 
\KwSty{procedure} \FuncSty{Simulate(}
                  $\mD, \mathcal{P}, d$\FuncSty{)} \;
\Indp
\lIf{$\gamma^{d} Rmax < \epsilon$}{
	\Return $0$\;
}
$s \leftarrow $ last state in $\mD$\;
\If{$N(\mD) = 0$}{
	\For{\KwSty{all} $a \in A$}{
		$N(\mD, a) \leftarrow 0$, $Q(\mD,a) \leftarrow 0$\;
	}
	$a \sim \pi_{ro}(\mD, \cdot)$\;
	$s' \sim \mathcal{P}(s,a,\cdot)$\;
	$r \leftarrow \mathcal{R}(s,a)$\;
	$R \leftarrow r + \gamma$ \FuncSty{Rollout(} $\mD\!as', \mathcal{P}, d$\FuncSty{)}\;	
	$N(\mD) \leftarrow 1$, $N(\mD, a) \leftarrow 1$\;
	$Q(\mD, a) \leftarrow R$\;
	\Return  $R$\;
}
$a \leftarrow \displaystyle\argmax_b Q( \mD, b) + c \sqrt{\tfrac{\log(N(\mD))}{N(\mD, b)}}$\;
$s' \sim \mathcal{P}(s,a,\cdot)$\;
$r \leftarrow \mathcal{R}(s,a)$\;
$R \leftarrow r + \gamma $ \FuncSty{Simulate(}$\mD\!as', \mathcal{P}, d\!+\!1$\FuncSty{)} \;
$N(\mD) \leftarrow N(\mD) + 1$\;
$N(\mD, a) \leftarrow N(\mD, a) + 1$\;
$Q(\mD, a) \leftarrow Q(\mD, a) + \frac{R-Q(\mD,a)}{N(\mD, a)}$\;
\Return $R$\;
\Indm
\KwSty{end procedure}\;
\end{algorithm2e}
\end{figure}
\begin{figure}[htbp]
\removelatexerror
\begin{algorithm2e}[H]
\caption{Thompson Sampling}
\label{alg:ts}
\DontPrintSemicolon

\KwSty{procedure} \FuncSty{GetAction(} $\mD$ \FuncSty{)} \;
\Indp
  $\mathcal{P} \sim P(\mathcal{P} | \mD)$ \;
  $\pi^* \leftarrow $ SolveMDP($\mathcal{P},\mathcal{R},\gamma$)\;
  $s \leftarrow $ last state in $\mD$ (current state)\;
  \Return $\pi^*(s)$\;
\Indm
\KwSty{end procedure}\;
\end{algorithm2e}
\end{figure}

\vspace{-0.15in}
\section{CRP mixture of {\sc mdp}s}
\label{sec:crpmdp}

To illustrate that our methodology is not restricted to tasks resembling
contextual bandits, we consider an extension of the problem in
Section~\ref{sec:crpcb} in which each subtask $\tau$ is a random {\sc
mdp} of a particular class, with contextual information being
informative about the class.

This extension is loosely motivated by the following oil exploration
problem. Possible drilling sites are considered in a sequence; each site
comes with some particular known contextual information (e.g.,
geological features). One can ignore a drilling site (action
$a_\textit{exit}$), or (buy acreage and) run one of two types of terrain
preparation strategies (actions $a_0$ and $a_1$ in state $s_0$). The
outcomes of these two strategies is the potential for either natural gas
(state $s_1$) or crude oil (state $s_2$). Finally, one has to make a
final choice about how to exploit (e.g., the type of extraction process,
either $a_2$ or $a_3$), which results in a stochastic payoff for this
drilling site: either dry hole with many expenses ($r=-1.5$) or a
profitable exploitation ($r=1$).

\begin{figure}[htpb]
\begin{center}
  \includegraphics[width=.33\textwidth]{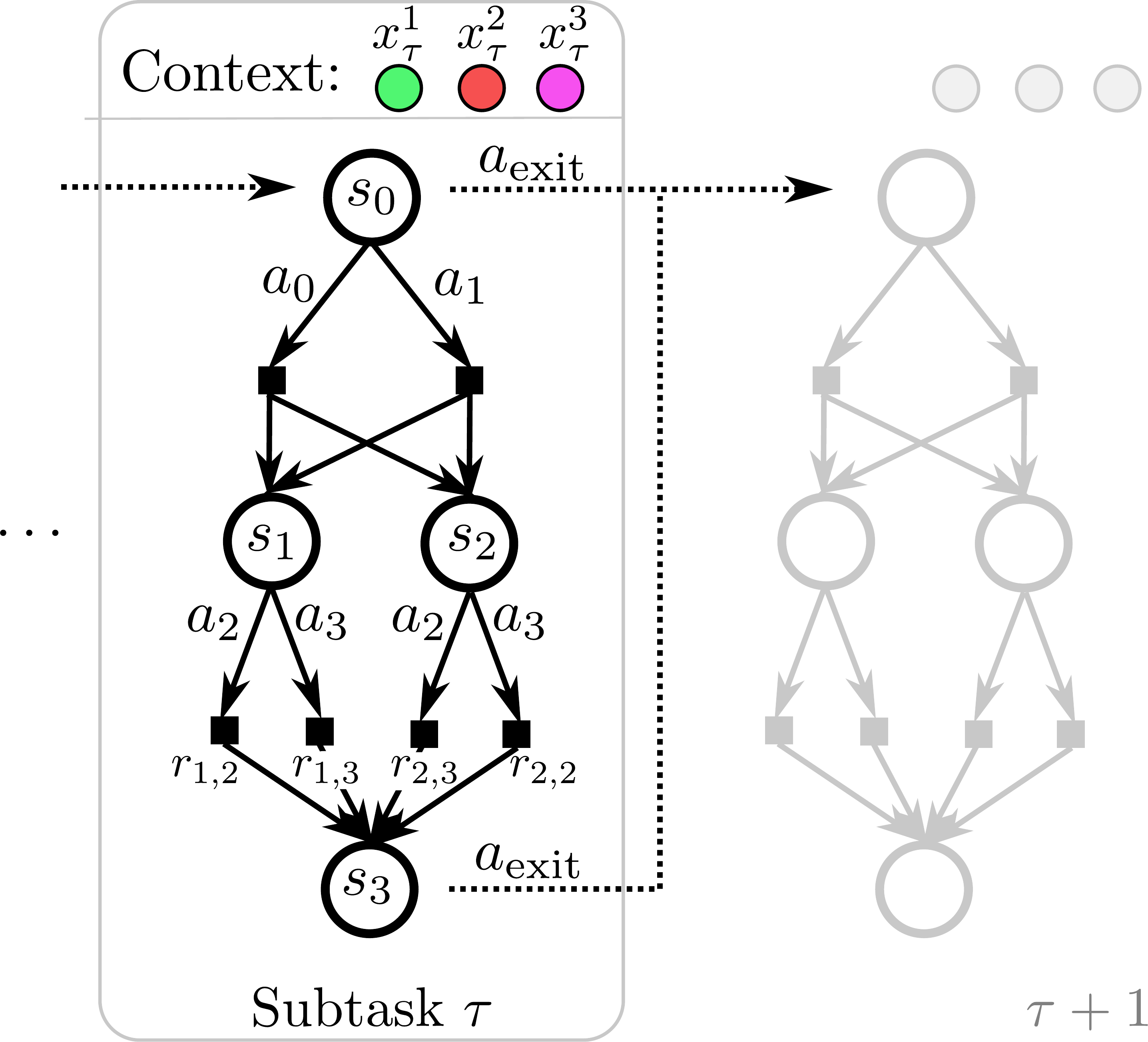}
\vspace{-0.1in}
\caption{The extension of the CRP mixture model to {\sc mdp}s, modelling
an intermediate decision within each subtask before getting a payoff.}
\label{fig:crpm}
\end{center}
\end{figure}

\begin{figure}[htpb]
  \begin{center}
    \subfigure{
    {\small(a)}
    \includegraphics[width=.23\textwidth]{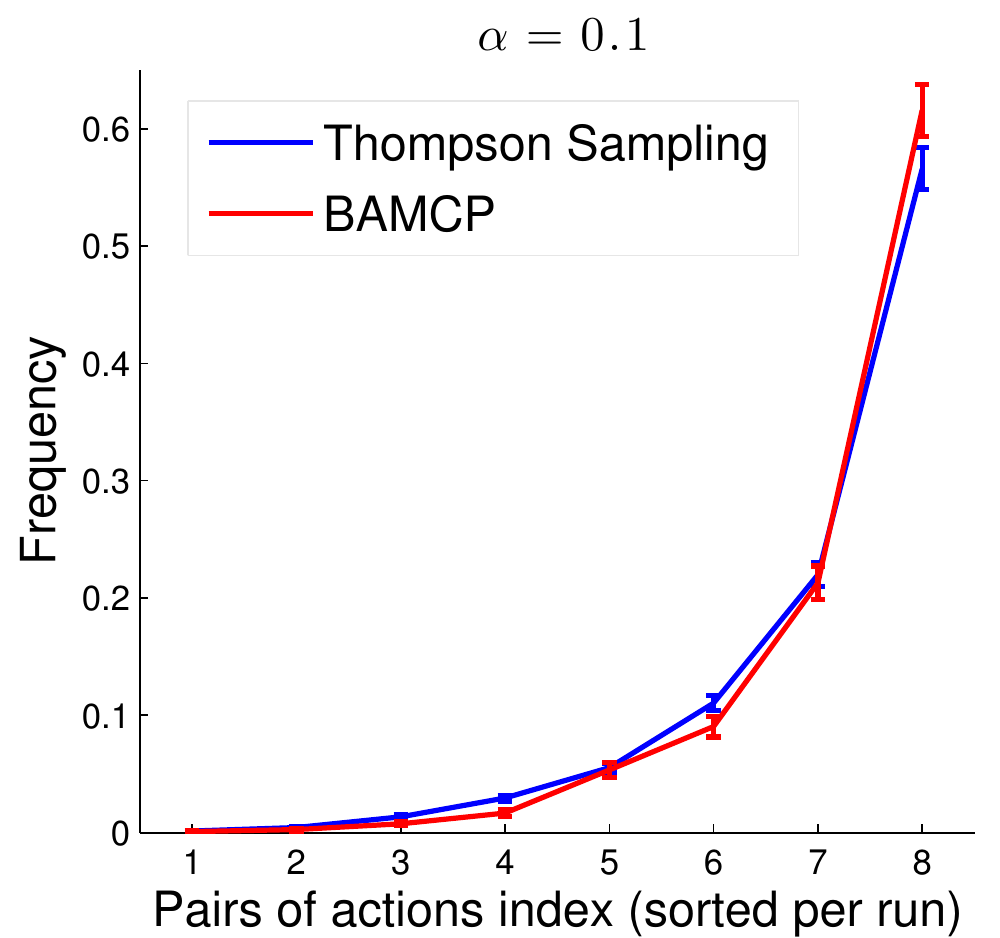}
    \includegraphics[width=.22\textwidth]{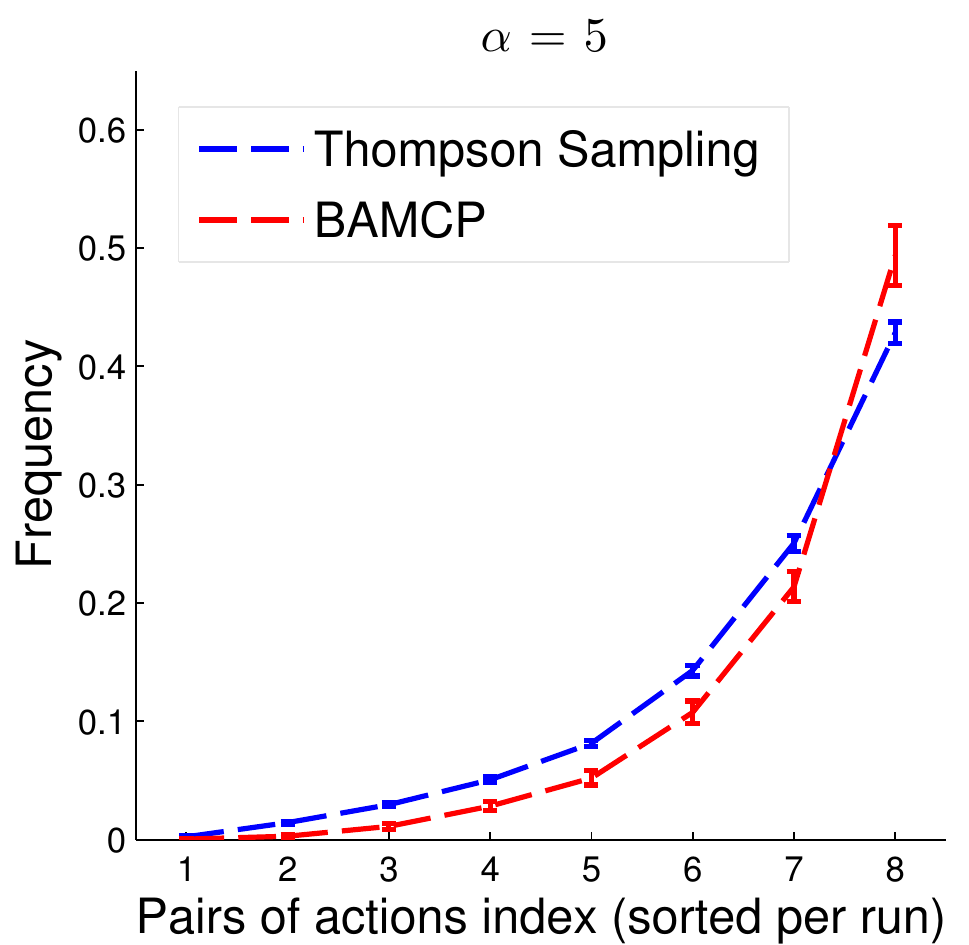}
    }
    \subfigure{
    {\small(b)}\includegraphics[width=.38\textwidth]{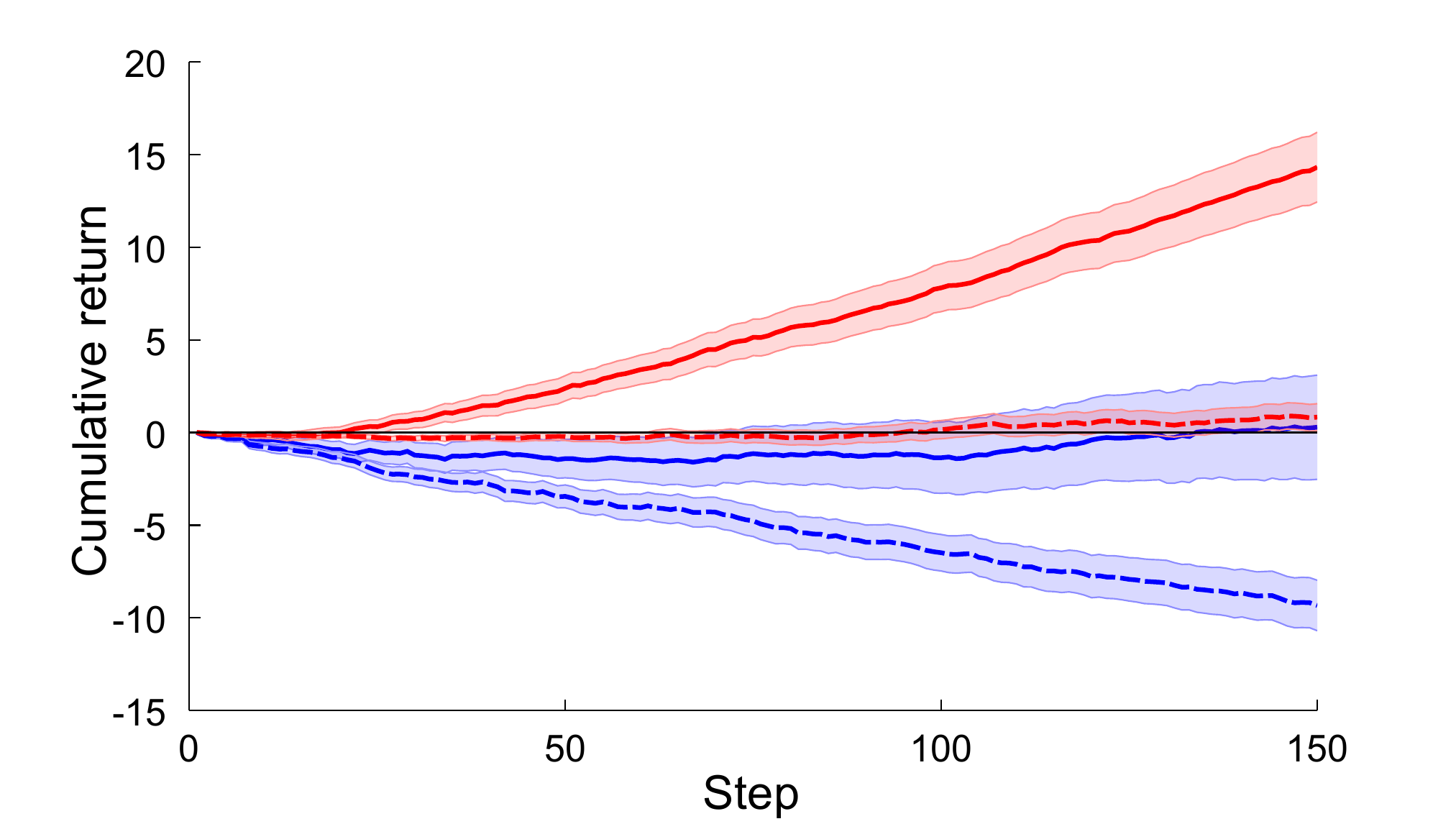}
    }
  \end{center}
  \vspace{-0.2in}
  \caption{\footnotesize{(a) Intratask exploration-exploitation statistics for
  {\sc bamcp} and {\sc ts} in the drilling problem. The distribution of action pairs
  (sorted in each run) executed {\em inside} the {\sc mdp} subtasks that were
  explored by the agent. Mean over 100 runs. (b) Comparison of
  the cumulative return at a fonction of the time step for {\sc bamcp}
  (red) and {\sc ts} (blue). Solid lines: $\alpha=0.1$, dotted line: $\alpha=5$. 
  }}
  \label{fig:crpmresults}
\end{figure}

We model this task as in Section~\ref{sec:crpcb} (each drilling site
corresponds to a cluster/subtask $\tau$), except that additional
$x_{\tau}$ variables are necessary to establish the transitions between
states, in addition to the variables already modeling context and
rewards (for states $s_1$ and $s_2$). The payoff $r_{1,3}$ from
executing action $a_3$ in $s_1$ in subtask $\tau$ is determined by the
binary variable $x_\tau^{1,3}$ as $r_{1,3} = f(x_\tau^{1,3})$
($f(0)=-1.5, f(1)=1$). In the generative model, $x_\tau^{1,3}$ (and
other $x_\tau^{i,j}$) is determined like the other $x_{\tau}$ variables
in Section~\ref{sec:crpcb}.  The model is illustrated in
Figure~\ref{fig:crpm}. The performance of {\sc bamcp} and {\sc ts} on
simulated data resemble that in the contextual bandits of
Section~\ref{sec:crpcb}. {\sc bamcp} adaptively ignores,
explores, or exploits drilling sites depending on the environmental
statistics. As in the other examples, with the same statistical model,
{\sc ts} acts too optimistically to do well in terms of discounted
return and is less inclined to ignore subtasks. Figure~\ref{fig:crpmresults}-a
shows that {\sc bamcp} is also more conservative when acting within each {\sc
mdp} when $\alpha$ is large, compared to {\sc ts}. The dynamics
of the cumulative return as a function of the steps is presented in 
Figure~\ref{fig:crpmresults}-b.

\end{document}